\documentclass[letterpaper, 10 pt, conference]{ieeeconf}  

\IEEEoverridecommandlockouts                              

\usepackage{amsmath} 
\usepackage{amssymb}  

\usepackage{adjustbox}
\usepackage{multirow}
\usepackage{wrapfig}
\usepackage{url}
\usepackage[bookmarks=true]{hyperref}
\usepackage[T1]{fontenc}
\usepackage{xcolor}

\title{\LARGE \bf
Adapting Pre-Trained Vision Models for Novel Instance Detection and Segmentation
}

\author{Yangxiao Lu, Jishnu Jaykumar P, Yunhui Guo, Nicholas Ruozzi, Yu Xiang
\thanks{All authors are with Department of Computer
Science, University of Texas at Dallas, Richardson, TX 75080, USA. {\tt\small \{yangxiao.lu, jishnu.p, yunhui.guo, nicholas.ruozzi, yu.xiang\}@utdallas.edu}
}
}

\begin{document}

\maketitle
\thispagestyle{empty}
\pagestyle{empty}



\begin{abstract}
  Novel Instance Detection and Segmentation (NIDS) aims at detecting and segmenting novel object instances given a few examples of each instance. We propose a unified, simple, yet effective framework (NIDS-Net) comprising object proposal generation, embedding creation for both instance templates and proposal regions, and embedding matching for instance label assignment. Leveraging recent advancements in large vision methods, we utilize Grounding DINO and Segment Anything Model (SAM) to obtain object proposals with accurate bounding boxes and masks. Central to our approach is the generation of high-quality instance embeddings. We utilized foreground feature averages of patch embeddings from the DINOv2 ViT backbone, followed by refinement through a weight adapter mechanism that we introduce. 
  
  We show experimentally that our weight adapter can adjust the embeddings locally within their feature space and effectively limit overfitting in the few-shot setting. Furthermore, the weight adapter optimizes weights to enhance the distinctiveness of instance embeddings during similarity computation. This methodology enables a straightforward matching strategy that results in significant performance gains. Our framework surpasses current state-of-the-art methods, demonstrating notable improvements in four detection datasets. In the segmentation tasks on seven core datasets of the BOP challenge, our method outperforms the leading published RGB methods and remains competitive with the best RGB-D method. We have also verified our method using real-world images from a Fetch robot and a RealSense camera.~\footnote{Project page with video, code and appendix: \url{https://irvlutd.github.io/NIDSNet/}}

%
  
\end{abstract}

\section{Introduction}

Novel Instance Detection and Segmentation (NIDS) is a crucial task in robot perception aimed at identifying and locating unseen instances in images or videos, given a few examples of each instance. Suppose that a robot needs to grasp a specific, novel object instance from a cluttered desk given only a small number of multi-view template images of the object. NIDS can provide the precise bounding box and mask of the target given a query image. 



The current paradigm for NIDS typically encompasses the following steps: (1) generating proposals from a query image, (2) obtaining embeddings of the proposals and the object templates, and (3) matching the embeddings of the proposals with those of the templates for identification. Recent work~\cite{li2024voxdet, shen2023instance, nguyen2023cnos, lin2023sam} has utilized various open-world detectors, e.g., the Segment Anything Model (SAM)~\cite{kirillov2023segment} or FastSAM~\cite{zhao2023fast}, to obtain object proposals. However, the use of open-world detectors often results in the generation of region-based proposals rather than actual \textit{object} proposals. This misidentification of regions as objects can lead to object identification issues. For example, a single object may be divided into multiple proposals, or some background regions may be misclassified as foreground objects. These false alarms cause issues for the following identification stage.

To embed proposals and templates, some existing works, e.g., \cite{shen2023instance, nguyen2023cnos,lin2023sam}, adopt the \(cls\) token of DINOv2~\cite{oquab2023dinov2, darcet2023vision}. \cite{li2024voxdet} used a 3D voxel representation. Ideally, for each specific unseen instance, the template embeddings from different views should be similar to each other, but markedly different from the embeddings of other instances. However, in these previous works, the embeddings of different instances may remain similar. 

\begin{figure}
    \centering
\includegraphics[width=\columnwidth]{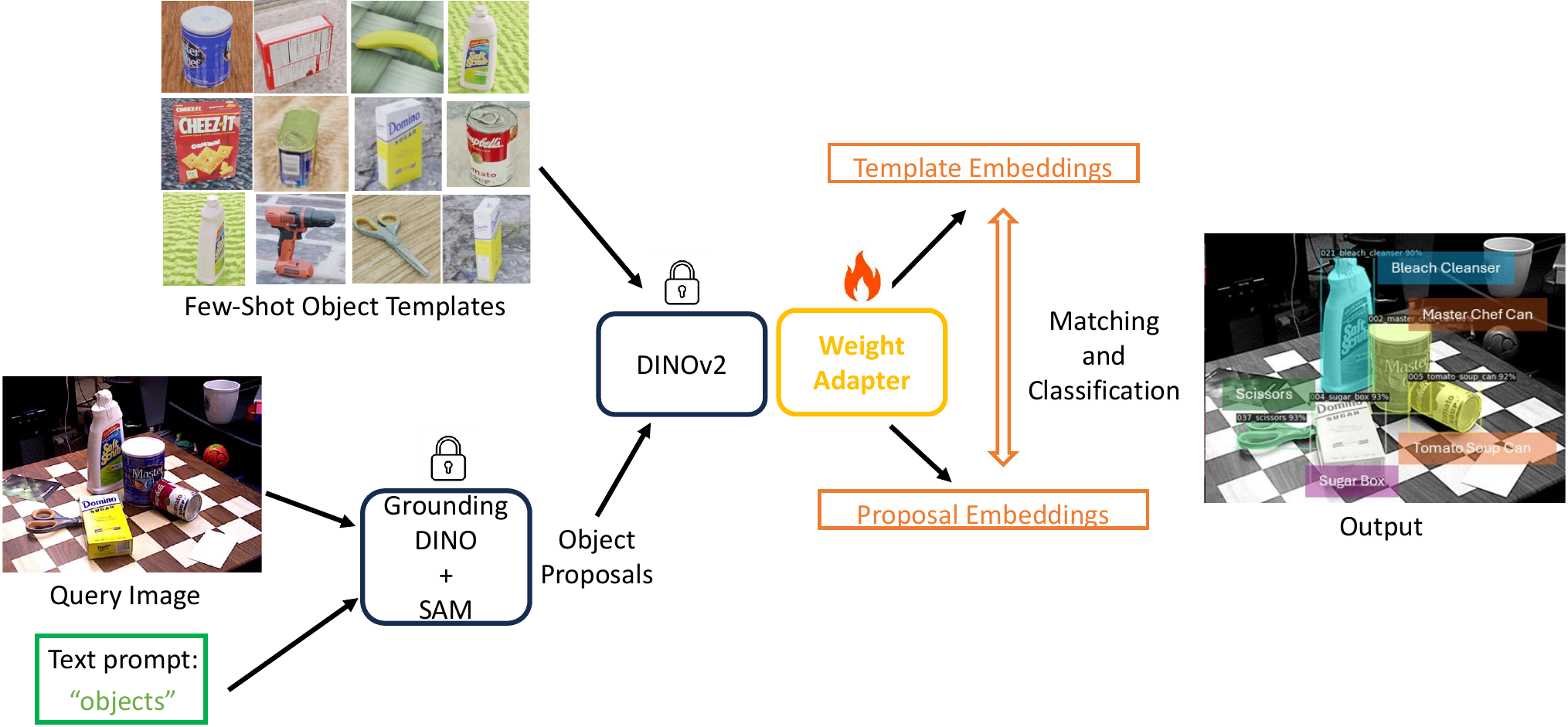}
    \caption{We leverage pre-trained vision models for object proposal generation and feature extraction, and introduce a weight adapter to improve pre-trained feature embeddings for novel object instance detection and segmentation.}
    \label{fig:pipeline}
\end{figure}

To address the limitations of existing methods, we propose a framework for NIDS, as depicted in Fig. \ref{fig:pipeline}. Initially, we utilize Grounding DINO~\cite{liu2023groundingDINO} with the text prompt \emph{``objects''} on a cluttered query image to obtain bounding boxes of foreground objects. This method capitalizes on the inherent objectness of Grounding DINO using specialized prompts. Subsequently, we employ SAM~\cite{kirillov2023segment} to create masks within these bounding boxes, generating precise object proposals comprising both bounding boxes and masks. For instance embeddings, we first extract average foreground features~\cite{kotar2024these} from the patch embeddings of the DINOv2 ViT backbone. We then apply adapters to enhance instance embeddings by clustering similar instances and distancing different ones. 

The adapters are trained using the template images only, where the InfoNCE loss~\cite{oord2018representation,chen2020simple} is applied to the feature embeddings after the adapters. To refine these embeddings, the CLIP-Adapter~\cite{gao2021clipAdapter} introduces a residual vector added to the original embedding, which risks overfitting with a few training examples. This addition can spoil and destabilize the embeddings of non-target objects, disrupting the entire feature space and causing the framework to misclassify non-targets as target instances. To mitigate this issue, we instead introduce a novel weight adapter (WA) that modifies the original embeddings by applying learned weights. Since the matching process relies on the standard cosine similarity, all embedding dimensions are treated equally, which may not sufficiently distinguish instances. The weights learned from our weight adapter emphasize the most relevant embedding channels. This produces more distinctive instance representations and improves performance by enhancing the discriminative capacity of embeddings. Finally, after applying the weight adapter to the embeddings, we employ a straightforward matching approach~\cite{mcvitie1971stable}, e.g., using stable matching or \(argmax\), to assign instance labels to these proposals from the query image.

Our approach has been validated on four detection datasets, seven segmentation datasets, and in the real world for NIDS. The framework significantly surpasses existing state-of-the-art methods. It demonstrates substantial increases in average precision (AP), with gains of 22.3, 46.2, 10.3, and 24.0 across these detection datasets. Moreover, our method outperforms the best published RGB and RGB-D methods on instance segmentation tasks over the seven core datasets of the BOP challenge~\cite{sundermeyer2023bop}.

Our contributions are summarized as follows.
\begin{itemize}
    \item We propose NIDS-Net, a unified framework for novel instance detection and segmentation that adapts pre-trained vision models for this task.
    \item We utilize the objectness of the Grounding DINO model and region recognition of the Segmentation Anything Model (SAM) to obtain object proposals.
    \item We introduce the Weight Adapter, a tool designed to refine embeddings within their feature space while preventing overfitting. This approach yields significant improvements as the adaptation of the feature space becomes more robust and stable. 
\end{itemize}

\section{Related Work}
\vspace{-1mm}

\textbf{Pretrained Models.} Large-scale pretraining has demonstrated utility across various downstream tasks. Works such as \cite{dosovitskiy2020ViT, caron2021dino, oquab2023dinov2} emphasize large-scale image representation learning. DINOv2~\cite{oquab2023dinov2} offers robust features to represent unseen instances. These foundational models offer diverse capabilities, with the challenge lying in effectively leveraging their wealth of knowledge for specific use cases, such as novel instance detection.  In our work, we leverage pre-trained vision models for NIDS.

\textbf{Instance Detection.}
Instance detection identifies an unseen instance in a test image using corresponding templates. Some methods such as~\cite{mercier2021deepTemplate, ammirato2018targetInstDet, liu2022gen6d}, rely on pure 2D representations and matching techniques. However, these methods may struggle with variations in 2D appearance caused by occlusion or pose changes. In contrast, VoxDet~\cite{li2024voxdet} utilizes explicit 3D knowledge from multi-view templates, providing geometry-invariant representations. We generate 2D robust instance embeddings from templates with DINOv2~\cite{oquab2023dinov2}. 


\textbf{Adapters for Pre-trained Models}. The application of adapters atop large pre-trained models has emerged as a prominent strategy, yielding significant improvements across various tasks. Previous works have leveraged adapters to enhance few-shot image classification tasks~\cite{gao2021clipAdapter, tip_adapter_eccv22, padalunkal2023protoclip}. For novel instance detection and segmentation, we propose the Weight Adapter to enhance performance. The adapter assigns weights to the embeddings from frozen backbones. SENets~\cite{hu2018squeeze} employ the Squeeze-and-Excitation (SE) blocks to distribute weights. In contrast, our adapter is an independent model with a structural difference from the SE block. For instance, the SE block lacks a ReLU layer prior to the sigmoid function.

\vspace{-2mm}
\section{Method}

The NIDS task is to locate and label novel object instances within a query image, given a set of template images of these objects. We assume that each of the \(N\) target instances is represented by \(K\) template images \(I_T \in \mathbb{R}^{K \times 3 \times W \times H}\) and their corresponding segmentation masks. During inference, the output of each query image \(I_Q \in \mathbb{R}^{3 \times W \times H}\) provides the bounding boxes for detection and instance masks for segmentation of these instances. 

\subsection{Instance Embedding Generation Stage}
In our approach, an instance embedding summarizes the pixels of an instance. The objective of this stage is to generate initial instance embeddings. Each of the \(N\) instances has \(K\) multi-view template images and their ground truth segmentation masks from which we derive template embeddings \(E_T \in \mathbb{R}^{N \times K \times C}\), where \(C\) is the dimension of the embeddings. Given an image and its corresponding mask, we initially extract patch embeddings using the ViT backbone of DINOv2~\cite{oquab2023dinov2}, and subsequently obtain foreground features as specified by the mask. We then perform average pooling on these features. This process, termed Foreground Feature Averaging (FFA), is proposed by \cite{kotar2024these} to assess object similarity. We employ FFA to generate all initial instance embeddings using the object templates.

\begin{figure*}
    \centering
\includegraphics[width=1.6\columnwidth]{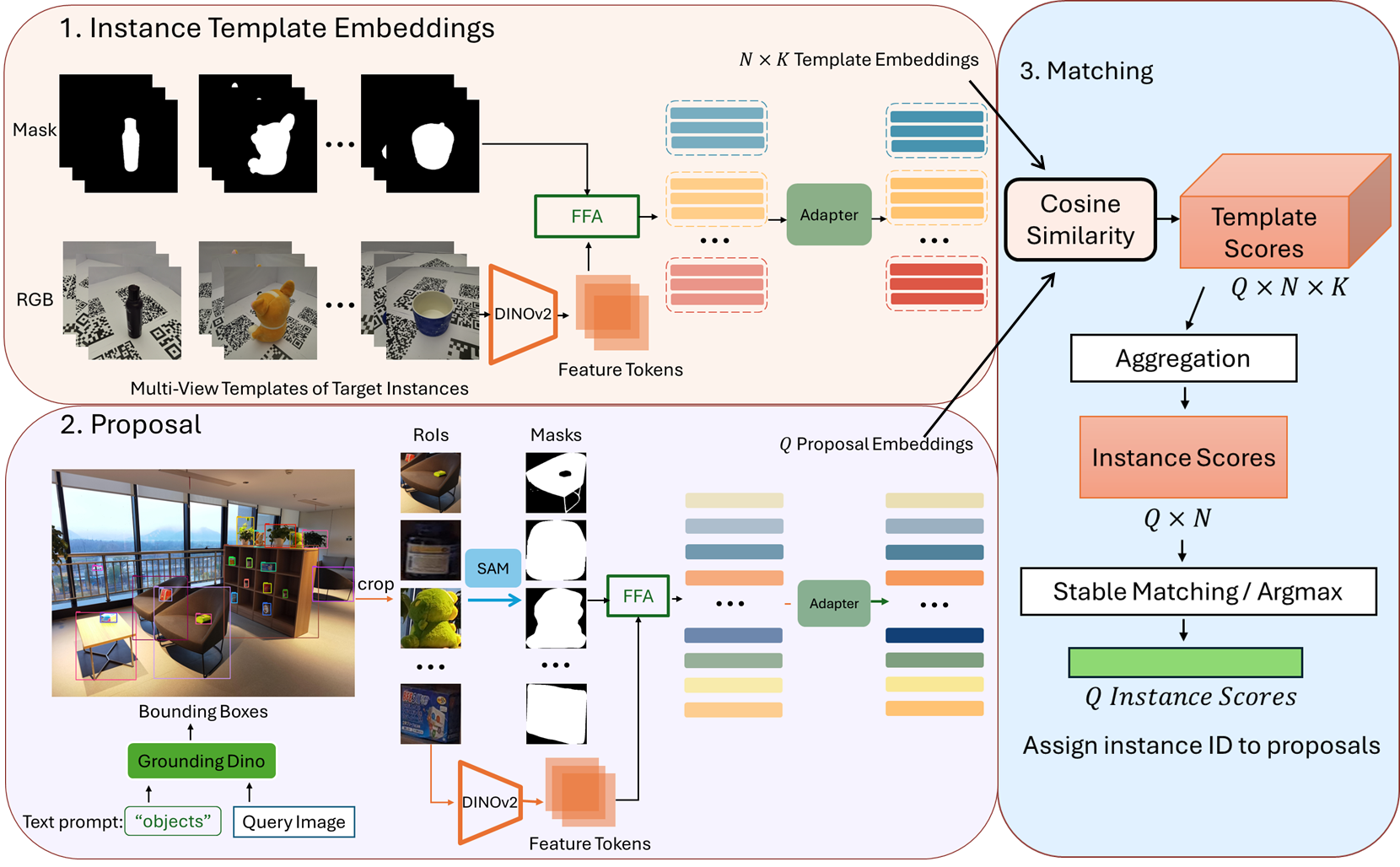}
    \caption{In our framework NIDS-Net, only adapters are learnable, while other models are frozen. Instance IDs are the instance labels.}
    \label{fig:framework}
    \vspace{-5mm}
\end{figure*}

\vspace{-1mm}
\subsection{Object Proposal Stage}
 \vspace{-1mm}
The purpose of this stage is to acquire object proposals from a query image. Previous studies~\cite{shen2023instance, nguyen2023cnos, lin2023sam} have utilized SAM~\cite{kirillov2023segment} or FastSAM~\cite{zhao2023fast} to generate regions as object proposals. Such proposals contain a high number of false alarms. For example, SAM might misclassify background regions or parts of objects as complete objects. To address this challenge, an off-the-shelf zero-shot detector, Grounding Dino~\cite{liu2023groundingDINO}, is employed with the text prompt ``objects'' to obtain initial bounding boxes of foreground objects. Then, SAM is applied to create masks based on these bounding boxes. The integration of these two models, termed Grounded-SAM (GS)~\cite{ren2024grounded}, significantly reduces the number of erroneous object proposals and expedites the subsequent stages. Moreover, this approach eliminates the need for training of the detector. We then extract the proposal regions along with their corresponding masks from the query images. Using the FFA pipeline with proposals, we calculate proposal embeddings \(E_P \in \mathbb{R}^{Q \times C}\), which correspond to \(Q\) regions of interest. Fig. \ref{fig:framework} illustrates the process of obtaining proposal embeddings.



\vspace{-2mm}
\subsection{Embedding Refinement via an Adapter}
\vspace{-1mm}
Most instance embeddings are well separated. However, some embeddings from distinct instances may cluster together. For example, non-target object embeddings might be similar to targets. Some target instances may resemble each other. To address this issue, we employ learnable adapters to refine the embeddings. We train these adapters using the InfoNCE loss~\cite{oord2018representation, chen2020simple}, aiming to bring the embeddings of the same instance closer together while separating the embeddings of different instances. \emph{This training only uses the few-shot template images.}

\textbf{CLIP-Adapter (CA)}. CLIP-Adapter~\cite{gao2021clipAdapter} comprises two trainable linear bottleneck layers appended to the language and image branches of CLIP. During few-shot fine-tuning, the original CLIP backbone remains frozen. Nonetheless, adding extra layers for fine-tuning may result in overfitting to the few examples available. To mitigate this issue, CLIP-Adapter integrates residual connections that dynamically merge the newly adapted knowledge with the foundational knowledge from the original CLIP backbone.

\begin{figure}
    \centering
\includegraphics[width=0.65\columnwidth]{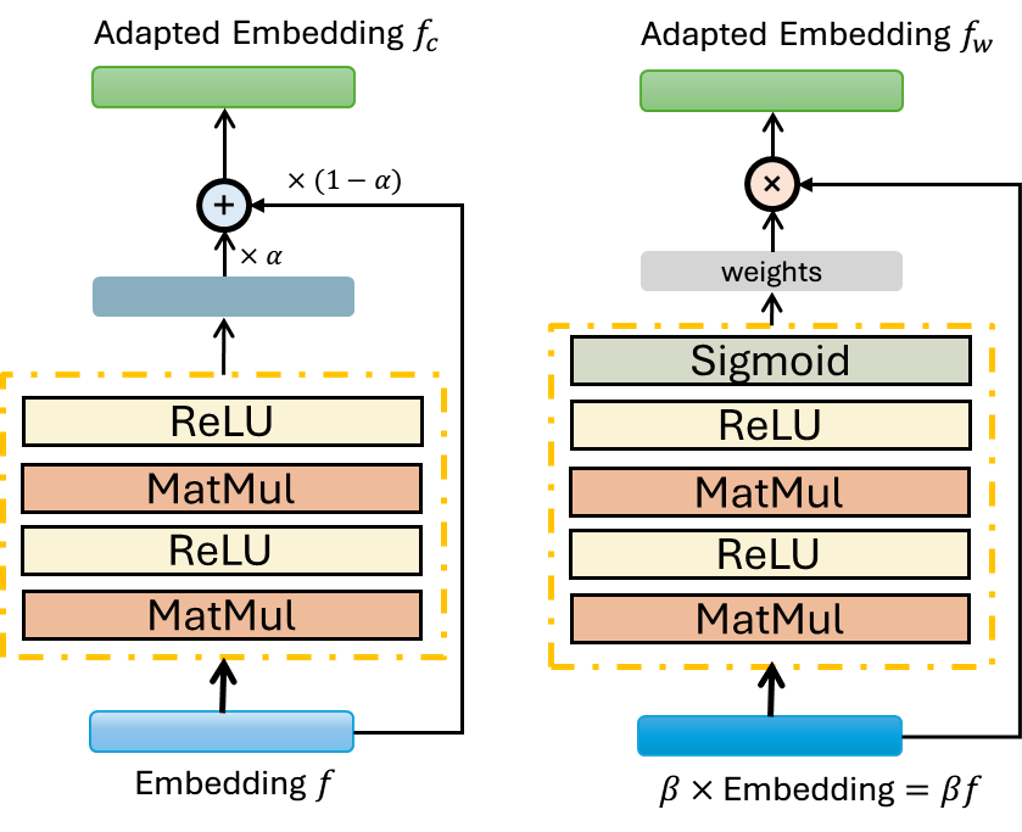}
\vspace{-3mm}
    \caption{(Left) CLIP-Adapter~\cite{gao2021clipAdapter} (Right) Our introduced weight adapter}
    \label{fig:adapters}
    \vspace{-5mm}
\end{figure}

As illusrated in Fig.~\ref{fig:adapters} (Left), given an instance embedding \(\textbf{f}\) and an \(\text{MLP}\) (Multi-Layer Perceptron), the CLIP-Adapter modifies the embedding according to the formula \(\label{clip}
    \mathbf{f_{c}} = \alpha \times \text{MLP}(\mathbf{f}) + (1-\alpha) \times \mathbf{f}\). In this equation, \(\mathbf{f_{c}}\) represents the adapted embedding, which is a linear combination of the transformed embedding \(\text{MLP}(\mathbf{f})\) and the original embedding \(\mathbf{f}\). The residual ratio \(\alpha\) is set as 0.6.

\textbf{Our Weight Adapter (WA)}. CLIP-Adapter refines instance embeddings by adding a new feature vector. However, this vector may not align with the original feature space, potentially causing overfitting. In datasets where only a few hundred template embeddings of target instances are refined, this adaptation can spoil non-target object embeddings by altering the relative distances between target and non-target instances. It leads the framework to misclassify non-targets as targets. Given the robustness and effectiveness of the original embedding space, it is essential to fine-tune instance embeddings within this space by constraining the adaptation. This can be achieved by applying weights to the original embeddings.

We propose the Weight Adapter, a compact MLP-based network structure illustrated in Fig.~\ref{fig:adapters} (Right). This adapter operates according to the following equations:
\vspace{-1mm}
\begin{align}
    \mathbf{w} &= \text{sigmoid}(\text{MLP}(\beta \mathbf{f})), \\
    \mathbf{f_{w}} &= \mathbf{w} \odot (\beta \mathbf{f}),
\end{align}
where $\beta$ is a predefined positive scalar. $\mathbf{w}$ represents the weights derived by passing the scaled embedding $\beta \mathbf{f}$ through an MLP, followed by a sigmoid activation function. The resultant $\mathbf{f_{w}}$ is the element-wise multiplication of $\mathbf{w}$ and $\beta \mathbf{f}$, yielding the weighted embedding. An ReLU activation precedes the sigmoid in our architecture, restricting the output to the range [0.5, 1), which ensures that each adapted embedding retains proximity to its original value.

We will use the cosine similarity between two embeddings to measure their similarity. For two embeddings $\mathbf{q}$ and $\mathbf{k}$, their cosine similarity is defined as:
\begin{equation}
    \cos(\mathbf{q}, \mathbf{k}) = \frac{\mathbf{q} \cdot \mathbf{k}}{\|\mathbf{q}\|\;\|\mathbf{k}\|}.
\end{equation}
With our weight adapter, we modify the embeddings using respective weight vectors $\mathbf{w}_1$ and $\mathbf{w}_2$ as follows:
\begin{align}
\mathbf{q}' = \mathbf{w}_1 \odot (\beta\mathbf{q}), \\
\quad \mathbf{k}' = \mathbf{w}_2 \odot (\beta\mathbf{k}).
\end{align}
Consequently, the cosine similarity between $\mathbf{q}'$ and $\mathbf{k}'$ is expressed as:
\begin{equation}
    \cos(\mathbf{q}', \mathbf{k}') = \frac{\sum_i w_{1,i}\,w_{2,i}\,q_i\,k_i}{\sqrt{\sum_i w_{1,i}^2 q_i^2}\;\sqrt{\sum_i w_{2,i}^2 k_i^2}}.
\end{equation}

Here, $\beta$ acts as a scaling factor since the embedding values can be small. We simply set $\beta$ to 10. It facilitates feature scaling, stabilizing gradients and enhancing convergence. The computed weights from the adapter, $\mathbf{w}_1$ and $\mathbf{w}_2$, ensure that embeddings of the same instance are closely aligned. These weights allow us to find the important dimensions for similarity computation. Moreover, our adapter is versatile and flexible because it can be integrated with any image encoder or embedding generation mechanism.


%


\vspace{-2mm}
\subsection{Matching Stage} 
This stage provides each proposal \(i\) with an instance ID \(o_i\) as its label and a confidence score \(s_i\). Initially, \(Q\) proposal embeddings \(E_P\) are matched with \(N \times K\) template embeddings \(E_T\) using cosine similarity. It yields a matrix of template scores with dimensions \(Q \times N \times K\), as illustrated in Fig. \ref{fig:framework}. For each instance, we aggregate all \(K\) template scores to derive a matrix of instance scores with dimensions \(Q \times N\). We employ \(Max\) as the aggregation function for optimal results.

\textbf{Bonus Instance Score for Segmentation.} To improve segmentation performance, we incorporate an additional appearance matching score, $s_{\text{appe}}$, as proposed by SAM-6D~\cite{lin2023sam}, into the instance scores. The final instance scores are computed as the average of $s_{\text{appe}}$ and the initial instance scores. \(s_{\text{appe}}\) is used to identify objects that are semantically similar yet differ in appearance. For each proposal, we can identify its most similar template \(T_{\text{best}}\) according to template scores. \(s_{\text{appe}}\) is derived from the patch embeddings of a proposal image \(I\) and \(T_{\text{best}}\). It quantifies the similarity between the query image and the best template in terms of their local features (\(f_{I, j}^{\text{patch}}\), \(f_{T_{\text{best}}, i}^{\text{patch}}\)) as follows:
\begin{equation}
s_{\text{appe}} = \frac{1}{N_{I}^{\text{patch}}} \sum_{j=1}^{N_{I}^{\text{patch}}} \max_{i=1,\ldots,N_{T_{\text{best}}}^{\text{patch}}} \left(\frac{f_{I, j}^{\text{patch}} \cdot f_{T_{\text{best}}, i}^{\text{patch}}}{\|f_{I, j}^{\text{patch}}\|_2 \|f_{T_{\text{best}}, i}^{\text{patch}}\|_2} \right),
\end{equation}
where \(N^{\text{patch}}\) is the number of patch embeddings. 

Assuming that objects are unique within a query image~\cite{shen2023instance}, we employ the stable matching algorithm~\cite{mcvitie1971stable} on instance scores to assign a unique instance ID to each proposal. If the assumption of uniqueness is not met, we use the \(Argmax\) function on the instance score matrix, which permits multiple proposals to share the same instance ID.

After matching, we acquire labeled proposals. Each is defined by \(\{b_i, M_i, o_i, s_i\}\). Here, \(b_i\) is the bounding box of an instance. \(M_i\) is the modal mask, which covers the visible object surface~\cite{sundermeyer2023bop}. \(o_i\) is the instance ID, and \(s_i\) is the confidence score. A confidence score threshold \(\delta\) can be used to remove incorrect proposals.


\begin{table*}[ht]
\centering
\caption{Detection results on the high-resolution dataset. \textit{avg} indicates that we are evaluating all images, including those from the easy and hard scenes. We train these two adapters with 2400 template embeddings. }
\label{tab:HR}
\vspace{-3mm}
{\scriptsize
\begin{tabular}{lcccccccc|c}
\hline
   \multirow{2}{*}{Method}    & \multicolumn{6}{c}{\textbf{AP}}                                        & \textbf{AP$_{50}$} & \textbf{AP$_{75}$} & \multirow{2}{*}{Time (sec)}\\ \cline{2-7}
           & avg                           & hard  & easy  & small & medium & large &               &               \\ \hline
FasterRCNN~\cite{ren2015faster} & 19.5 & 10.3 & 23.8 & 5.0 & 22.2 & 38.0 & 29.2 & 23.3  & 0.00399  \\
RetinaNet~\cite{lin2017focal}  & 22.2 & 14.9 & 26.5 & 5.5 & 25.8 & 42.7 & 31.2 & 25.0 & 0.00412   \\
CenterNet~\cite{zhou2019objects}  & 21.1 & 11.8 & 25.7 & 5.9 & 24.1 & 40.4 & 32.7 & 23.6   & 0.00376      \\
FCOS~\cite{9010746}      & 22.4 & 13.2 & 28.7 & 6.2 & 26.5 & 38.1 & 32.8 & 25.5 & 0.00271 \\
DINO~\cite{zhang2022dinodetr}      & 28.0 & 17.9 & 32.6 & 11.5 & 30.7 & 45.1 & 32.2 & 32.2  & 1.90625 \\
SAM + DINO$_{f}$~\cite{caron2021emerging}  & 37.0 & 22.4 & 43.9 & 11.9 & 40.9 & 62.7 & 44.1 & 40.4  & 15.10 \\
SAM + DINOv2$_{f}$~\cite{oquab2023dinov2}  & 41.6 & 28.0 & 47.6 & 14.6 & 45.8 & 69.1 & 49.1 & 46.0 & 14.70 \\
\hline
NIDS-Net w/o adapter (Ours) & 59.3  & 41.7 & 67.4  &   12.7  & 58.1   & 78.8  & 71.1    & 65.1 
 & 6.92 \\
NIDS-Net + CA (Ours)  & 61.3 & 42.5 & 69.6 &  17.0 & 59.3 & 81.6 & 73.2 & 67.0 & 6.99 \\
NIDS-Net + WA (Ours)  & \textbf{63.9} & \textbf{43.4} & \textbf{72.7} &  \textbf{18.1} & \textbf{62.5} & \textbf{84.0} & \textbf{76.6} & \textbf{70.6} & 6.78\\ 
\hline
\end{tabular}
}
\vspace{-5mm}
\end{table*}

\section{Experiments}
We employ the Grounding DINO model with a Swin-T backbone~\cite{liu2023groundingDINO} and a text prompt ``objects'', and the default ViT-H SAM~\cite{kirillov2023segment} to generate object proposals. According to the proposals, we obtain the regions of interest (RoIs) and resize them to \(448 \times 448\) or \(224 \times 224\) resolutions. For object detection, instance embeddings are produced using the DINOv2's ViT-L model with registers~\cite{darcet2023vision}, employing stable matching due to the uniqueness of instances in these datasets. For image segmentation,  following SAM-6D~\cite{lin2023sam}, we utilize the ViT-L model of DINOv2~\cite{oquab2023dinov2}. In scenarios with identical instances, such as the cluttered scenes in BOP datasets, we apply the \(argmax\) function for matching.

The experiments are conducted on an RTX A5000 GPU. For each dataset, we train both adapters on template embeddings using the InfoNCE loss~\cite{oord2018representation, chen2020simple}. 
With two linear layers, both adapters initially decrease the embedding dimension from \(C\) to \(C/4\), subsequently restoring it to \(C\). The Weight Adapter is trained with the Adam optimizer~\cite{kingma2014adam} at a learning rate of 1e-3 and a batch size of 1024, while the CLIP-Adapter is trained at a learning rate of 1e-4. Dropout~\cite{hinton2012improving} is incorporated in the CLIP-Adapter to reduce overfitting. For additional details, please see the appendix.

\vspace{-1mm}
\subsection{Detection Datasets}
\vspace{-1mm}
We utilize recently developed instance detection datasets and their associated baselines \cite{li2024voxdet, shen2023instance}. The High-resolution and RoboTools datasets employ real template images, while the LM-O and YCB-V datasets utilize synthetic images.

\textbf{High-resolution Dataset.} \cite{shen2023instance} design a high-resolution dataset for instance detection. This dataset comprises 100 distinct object instances, each represented by 24 photos taken from multiple viewpoints, with each photo having a resolution of \(3072 \times 3072\) pixels. These instances are integrated into 14 different indoor scenes for testing, captured in even higher resolution (\(6144 \times 8192\) pixels). The testing images are further classified as easy or hard based on the level of scene clutter and how much the instances are occluded. Easy tags are assigned when objects are sparsely placed, while hard tags are used for more cluttered setups. 

According to \cite{shen2023instance}, two types of baselines are set for the high-resolution dataset: Cut-Paste-Learn and a non-learned method. Cut-Paste-Learn generates synthetic training images with 2D-box annotations, putting foreground instances in various sizes and aspect ratios onto different backgrounds. This enables training detectors by viewing each instance as a class. They train five detectors: FasterRCNN~\cite{ren2015faster}, RetinaNet~\cite{lin2017focal}, CenterNet~\cite{zhou2019objects}, FCOS~\cite{9010746}, and the transformer-based DINO~\cite{zhang2022dinodetr}. For the non-learned method, SAM is initially used to generate proposals, followed by employing DINO~\cite{caron2021emerging} and DINOv2~\cite{oquab2023dinov2} to generate features for both proposal and template images, ultimately performing proposal matching and selection.

\textbf{Synthetic-Real Test Sets.} \cite{li2024voxdet} employ two benchmarks for evaluation. LineMod-Occlusion (LM-O)~\cite{brachmann2014learning} includes 8 texture-less objects and 1,514 bounding boxes. The YCB-Video (YCB-V)~\cite{calli2015ycb} features 21 objects and 4,125 bounding boxes. Since these datasets have real testing images without reference videos, \cite{li2024voxdet} generate synthetic videos using CAD models. We sample 16 synthetic template images per object from these videos. 


\textbf{RoboTools Benchmark}~\cite{li2024voxdet} features 20 distinct instances, 9,109 annotations, and 24 complex scenarios. 25 real template images per instance are sampled from their reference videos. For cluttered scenes from the datasets RoboTools, LM-O and YCB-V, \cite{li2024voxdet} have developed several 2D baselines: OLN$_{DINO}$, OLN$_{CLIP}$, and OLN$_{Corr}$. Initially, they generate open-world 2D proposals using their detection module~\cite{kim2022learning}. For matching, different methods are employed to select the proposal with the highest score. In OLN$_{DINO}$ and OLN$_{CLIP}$, they use robust features from pre-trained backbones~\cite{caron2021emerging, radford2021learning} and cosine similarity for matching. For OLNCorr, a matching head is designed based on correlation~\cite{liu2022gen6d}. They also employ the class-level one-shot detectors OS2D~\cite{osokin2020os2d} and BHRL~\cite{yang2022balanced}. To address the limitations of traditional 2D methods with pose variations and occlusions, \cite{li2024voxdet} introduced VoxDet which is based on a 3D voxel representation. To train these methods, \cite{li2024voxdet} developed a synthetic instance detection dataset (OWID-10k). 


\textbf{Evaluation Metrics.} For detection, we evaluate our method with Average Precision (AP). AP is computed by averaging precision scores at various Intersection over Union (IoU) thresholds, specifically from 0.5 to 0.95, in increments of 0.05~\cite{lin2014microsoft}. Additionally, AP50 and AP75 are variations of this metric, where precision is averaged across all instances at IoU thresholds of 0.5 and 0.75, respectively. 

\subsection{Segmentation Datasets}

\textbf{The BOP Challenge.} We test our method using seven  datasets from the BOP challenge~\cite{sundermeyer2023bop}: LineMod Occlusion (LM-O)~\cite{brachmann2014learning}, T-LESS~\cite{hodan2017t}, TUD-L~\cite{hodan2018bop}, IC-BIN~\cite{doumanoglou2016recovering}, ITODD~\cite{drost2017introducing}, HomebrewedDB (HB)~\cite{kaskman2019homebreweddb}, and YCB-Video~\cite{calli2015ycb}. We use 42 template rendering images from CNOS~\cite{nguyen2023cnos}, which are generated via BlenderProc~\cite{denninger2019blenderproc}.

\textbf{Baselines.} We compare our method with ZeroPose~\cite{chen20233d}, CNOS~\cite{nguyen2023cnos}, and SAM-6D~\cite{lin2023sam}. These methods utilize proposals from SAM~\cite{kirillov2023segment} or FastSAM~\cite{zhao2023fast} and use the $cls$ token from DINOv2 as the instance embedding for matching. In addition to adjusting the SAM hyperparameters to generate more proposals, SAM-6D enhances its performance by incorporating an appearance score, \(s_{\text{appe}}\). Moreover, SAM-6D employs a Geometric Matching Score, \(s_{\text{geo}}\), which considers the shapes and sizes of instances during matching by utilizing depth information. 

\textbf{Evaluation Metrics.} For instance segmentation task, we evaluate our method using the Average Precision (AP) metrics. The AP is computed by averaging the precision scores at various IoU thresholds, from 0.50 to 0.95, increasing by 0.05 at each step.



\begin{table}
\centering
\caption{Detection performance on the fully real dataset, RoboTools. \textbf{Proposal} indicates the object proposal model. \textit{GS} stands for Grounded-SAM. OLN* is trained with the matching head while OLN employs fixed modules. }
\label{tab:robotools}
\vspace{-2mm}
{\scriptsize
\begin{tabular}{lccccc}
\hline
\textbf{Method} &
  \textbf{Proposal} &
  \textbf{AP} &
  \textbf{AP$_{50}$} &
  \textbf{AP$_{75}$} 
  \\ \hline
OS2D~\cite{osokin2020os2d} & N/A & 2.9  & 6.5  & 2.0  \\
DTOID~\cite{mercier2021deep} & N/A  & 3.6  & 9.0  & 2.0  \\
OLN\textsubscript{Corr.}~\cite{kim2022learning, liu2022gen6d} & OLN*  & 14.4 & 18.1 & 15.7 \\
VoxDet~\cite{li2024voxdet}   & OLN* & 18.7 & 23.6 & 20.5 \\
\hline
NIDS-Net w/o adapter (Ours)  & GS  & 63.3          & 77.2          & 68.7          &  \\
NIDS-Net + CA (Ours)  &   GS     & 64.8          & 79.2          & 70.7       &  \\
NIDS-Net + WA (Ours)     &  GS   & \textbf{64.9} & \textbf{79.4} & \textbf{70.8} &   \\
\hline
\end{tabular}
}
\vspace{-2mm}
\end{table}

\begin{table*}[h]
\centering
\caption{Detection performance on the LM-O and YCB-V datasets. OLN* is trained alongside the matching head whereas OLN uses fixed modules. \dag means the model is trained on OWID and real images. \((\cdot)\) represents the number of instances.} 
\label{tab:lmo}
{\scriptsize
\begin{tabular}{lcc|ccc|ccc|cccc}
\hline
\multicolumn{3}{c|}{}                               & \multicolumn{3}{c|}{LM-O (8)}    & \multicolumn{3}{c|}{YCB-V (21)} & \multicolumn{4}{c}{Average}   \\
Method                     & Proposal & Train                   & AP  & AP$_{50}$ & AP$_{75}$      & AP     & AP$_{50}$    & AP$_{75}$   & AP  & AP$_{50}$ & AP$_{75}$& Time (s) \\ \hline
OS2D~\cite{osokin2020os2d}             & N/A     & OWID       & 0.2  & 0.7  & \textless{}0.1 & 5.2     & 18.3    & 1.9    & 2.7  & 9.5  & 1.0  & 0.189   \\
DTOID~\cite{mercier2021deep}              & N/A     & OWID         & 9.8  & 28.9 & 3.7            & 16.3    & 48.8    & 42     & 13.1 & 38.9 & 4.0  & 0.357    \\
OLN\textsubscript{CLIP}~\cite{kim2022learning, radford2021learning}   & OLN     & OWID\dag & 16.2 & 32.1 & 15.3           & 10.7    & 25.4    & 7.3    & 13.5 & 28.8 & 11.3 & 0.357   \\
Gen6D~\cite{liu2022gen6d}          & N/A     & OWID\dag  & 12.0 & 29.8 & 6.6            & 8.1     & 33.0    & 5.5    & 18.4 & 33.5 & 5.9  & 0.769   \\
BHRL~\cite{yang2022balanced}            & N/A     & COCO                   & 14.1 & 21.0 & 15.7           & 31.8    & 47.0    & 34.8   & 23.0 & 34.0 & 25.3 & N/A   \\
OLN\textsubscript{Corr.}~\cite{kim2022learning, liu2022gen6d}    & OLN*    & OWID                    & 22.3 & 34.4 & 24.7           & 24.8    & 41.1    & 26.1   & 23.6 & 37.8 & 25.4 & 0.182   \\
OLN\textsubscript{Dino}~\cite{kim2022learning, caron2021emerging}     & OLN     & OWID\dag & 23.6 & 41.6 & 24.8           & 25.6    & 53.0    & 21.1   & 24.6 & 47.3 & 23.0 & 0.357   \\
VoxDet~\cite{li2024voxdet}   & OLN*    & OWID                    & 29.2 & 43.1 & 33.3           & 31.5    & 51.3    & 33.4   & 30.4 & 47.2 & 33.4 & 0.154   \\
 \hline
NIDS-Net w/o adapter (Ours)  &   GS    &  N/A    & 38.7          & 66.0 & 41.0          & 53.0    & 72.9    & 61.7   & 45.9 & 69.5 & 51.4 & 3.73    \\
NIDS-Net + CA (Ours)  &    GS    &  N/A   & 39.2 & 67.0 & 41.4 & 53.9 & 74.1 & 62.7 & 46.6 & 70.6 & 52.1 & 3.71  \\ 
NIDS-Net + WA (Ours)  &     GS    &  N/A      & \textbf{39.5} & \textbf{67.4} & \textbf{41.8} & \textbf{55.5} & \textbf{75.5} & \textbf{65.0} & \textbf{47.5} & \textbf{71.5} & \textbf{53.4} &  3.73 \\
\hline
\end{tabular}
}
\vspace{-3mm}
\end{table*}

\begin{table*}[h]
\centering
\caption{Novel instance segmentation results on the seven core datasets of the BOP benchmark. We utilize Average Precision (AP) to compare these methods. \((\cdot)\) includes the number of instances. SAM6D (\emph{RGBD}) exclusively utilizes RGB-D images, whereas other models employ only RGB images.}
\vspace{-3mm}
\label{tab:bop}
{\scriptsize
\begin{tabular}{lcccccccc}
\hline
\multirow{2}{*}{Method}   & \multicolumn{7}{c}{BOP Datasets}                      &      \\ \cline{2-8}
                          & LM-O (8) & T-LESS (30) & TUD-L (3) & IC-BIN (2) & ITODD (28) & HB (33)  & YCB-V (21) & Mean \\ \hline
ZeroPose (SAM)~\cite{chen20233d}            & 34.4 & 32.7   & 41.4  & 25.1   & 22.4  & 47.8 & 51.9  & 36.5 \\
CNOS (SAM)~\cite{nguyen2023cnos}                & 39.6 & 39.7   & 39.1  & 28.4   & 28.2  & 48.0 & 59.5  & 40.4 \\
CNOS (FastSAM)~\cite{nguyen2023cnos}            & 39.7 & 37.4   & 48.0  & 27.0   & 25.4  & 51.1 & 59.9  & 41.2 \\
SAM-6D (FastSAM)~\cite{lin2023sam} & 39.5 & 37.6 & 48.7 & 25.7 & 25.3 & 51.2 & 60.2 & 41.2 \\
SAM-6D (FastSAM) + \(s_{\text{appe}}\)~\cite{lin2023sam} & 40.6 & 39.3 & 50.1 & 27.7 & 29.0 & 52.2 & 60.6 & 42.8 \\
SAM-6D (SAM)~\cite{lin2023sam}       & 43.4 & 39.1   & 48.2  & 33.3   & 28.8  & 55.1 & 60.3  & 44.0 \\
SAM-6D (SAM) +\(s_{\text{appe}}\)~\cite{lin2023sam} & 44.4 & 40.8   & 49.8  & 34.5   & 30.0  & 55.7 & 59.5  & 45.0 \\
SAM-6D (SAM, \textit{RGBD})~\cite{lin2023sam}          & \textbf{46.0} & 45.1 & \textbf{56.9} & \textbf{35.7} & \textbf{33.2} & 59.3 & 60.5 & 48.1 \\ \hline
NIDS-Net w/o adapter (Ours)  & 42.9 & 43.0   & 52.0  & 30.5   & 28.8  & 56.6 & 59.7  & 44.8 \\
NIDS-Net + CA (Ours)    & 42.8 & 47.5 & 50.0 & 29.7 & 27.6 & \textbf{62.7} & 63.5 & 46.3 \\
NIDS-Net + WA (Ours)   & 43.2 & 49.2 & 53.5 & 30.5 & 29.4 & 62.4 & 64.3 & 47.5 \\
NIDS-Net + WA + \(s_{\text{appe}}\) (Ours) & 43.9 & \textbf{49.6} & 55.6 & 32.8 & 31.5 & 62.0 & \textbf{65.0} & \textbf{48.6} \\
\hline
\end{tabular}
}
\vspace{-2mm}
\end{table*}

\begin{table*}[t]
\centering
\vspace{-1mm}
\caption{Segmentation performance comparison across proposal models and embedding methods. The first row is from CNOS~\cite{nguyen2023cnos}. }
\vspace{-2mm}
\label{tab:gsam}
{\scriptsize
\begin{tabular}{cccccccccc}
\hline
\multirow{2}{*}{Proposal} & \multirow{2}{*}{Embedding} & \multicolumn{7}{c}{BOP Datasets} &  \\ \cline{3-9}
 &  & LM-O & T-LESS & TUD-L & IC-BIN & ITODD & HB & YCB-V & Mean \\ \hline
SAM & cls token & 39.6 & 39.7 & 39.1 & 28.4 & 28.2 & 48.0 & 59.5 & 40.4 \\
SAM & FFA & 40.0 & 40.0 & 40.8 & 28.0 & 25.8 & 46.0 & 54.4 & 39.3 \\
GS & cls token & 41.7 & 41.7 & 50.8 & \textbf{31.5} & 30 & 58 & 63 & 45.2 \\
GS & FFA & 42.9 & 43.0 & 52.0 & 30.5 & 28.8 & 56.6 & 59.7 & 44.8 \\
GS & cls token + WA & 42.5 & 48.2 & 50.7 & 31.3 & \textbf{32.0} & 60.5 & 62.8 & 46.9 \\
GS & FFA + WA & \textbf{43.2} & \textbf{49.2} & \textbf{53.5} & 30.5 & 29.4 & \textbf{62.4} & \textbf{64.3} & \textbf{47.5} \\ \hline
\end{tabular}
}
\vspace{-3mm}
\end{table*}

\begin{figure}
    \centering
 \vspace{-1mm}    
\includegraphics[width=0.9\columnwidth]{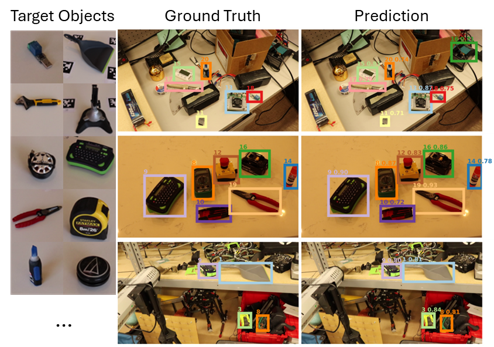}
\vspace{-2mm}
    \caption{Visual results on the RoboTools benchmark. }
    \label{fig:RT}
    \vspace{-5mm}
\end{figure}

\begin{figure}
    \centering
    \vspace{-2mm}
\includegraphics[width=0.9\columnwidth]{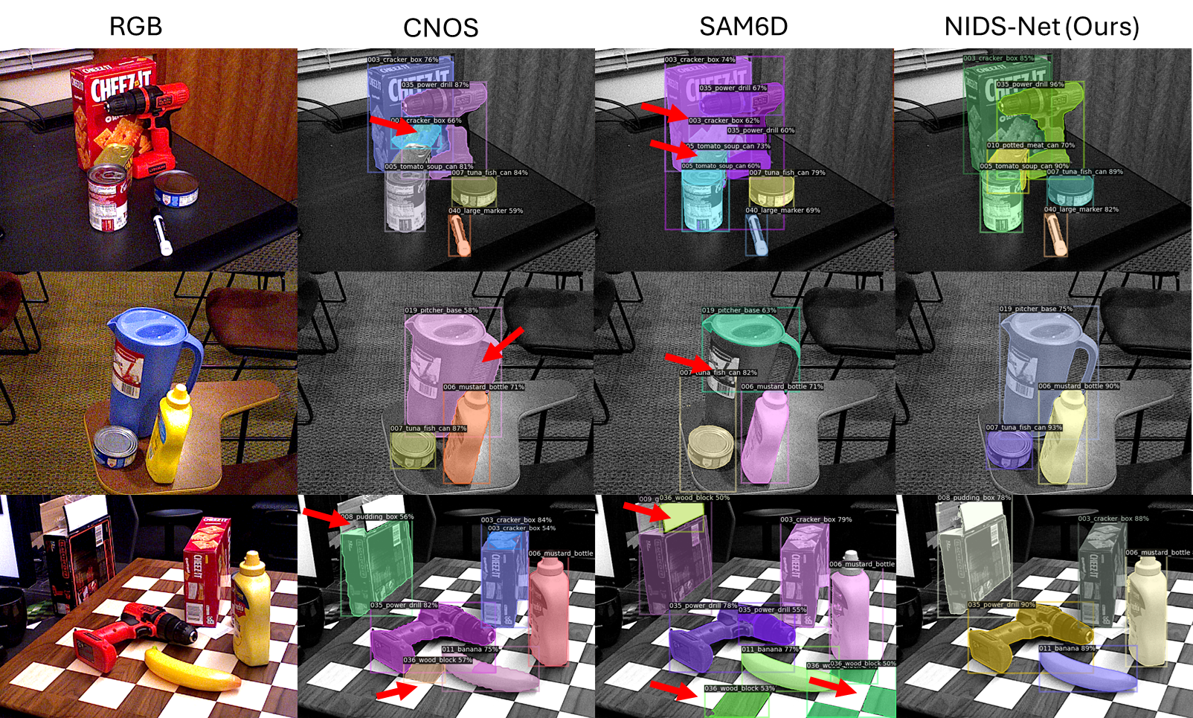}
    \caption{Comparison of segmentation results using CNOS, SAM6D, and NIDS-Net on the YCB-V dataset. CNOS and SAM6D may misclassify some background regions or object parts as objects due to proposal generation of SAM. Red arrows indicate these mistakes. NIDS-Net addresses this limitation with Grounded-SAM.}
    \vspace{-2mm}
    \label{fig:ycbv_comparison}
    \vspace{-5mm}
\end{figure}

\vspace{-1mm}
\subsection{Benchmarking Results}
\vspace{-1mm}
\textbf{Detection.} The high-resolution dataset results are presented in Table \ref{tab:HR}. Our models dramatically outperform existing techniques, achieving the highest AP scores across all categories. Our basic method surpasses the top baseline by 17.7 AP. Additionally, as this dataset contains 100 instances with some similarities among them, our Weight Adapter boosts overall performance by 4.6 AP and improves the detection of small objects by 5.4 AP. For the RoboTools dataset, detailed in Table \ref{tab:robotools}, our method outperforms the state-of-the-art VoxDet by 46.2 AP. Since RoboTools contains only 20 instances, our adapter has limited scope to enhance performance. Our approach achieves over 60 AP on these two fully real datasets. 

For the Synthetic-Real datasets, LM-O and YCB-V, we present their results in Table \ref{tab:lmo}. Our model outperforms VoxDet by 10.3 AP on LM-O and by 24.0 AP on YCB-V. The improved performance on YCB-V, which contains more instances, indicates that our adapter functions more effectively with increased instance variety.

\textbf{Segmentation.} CNOS~\cite{nguyen2023cnos} and SAM-6D~\cite{lin2023sam} initially utilize SAM or FastSAM to obtain proposals and derive instance embeddings through the \(cls\) token of Dinov2. In contrast, we generate proposals using Grounded-SAM and acquire embeddings through the FFA pipeline. The segmentation results for the seven principal datasets of the BOP challenge are detailed in Table \ref{tab:bop}. Overall, our method surpasses the state-of-the-art SAM-6D and achieves superior results compared to RGBD results. 

\textbf{Qualitative Results.} We present detection results of RoboTools in Fig. \ref{fig:RT}. Moreover, Fig. \ref{fig:ycbv_comparison} displays a visual comparison of the segmentation results. Additional examples are available in the appendix. 

\textbf{Runtime.} Table \ref{tab:HR} includes the runtime of our method for object detection. Grounded-SAM substantially decreases the number of proposals compared to SAM, thereby accelerating the entire process. Please see the appendix for segmentation runtime. 

\subsection{Real-world Testing and Failure Cases}

\begin{figure}
    \centering
\includegraphics[width=0.8\columnwidth]{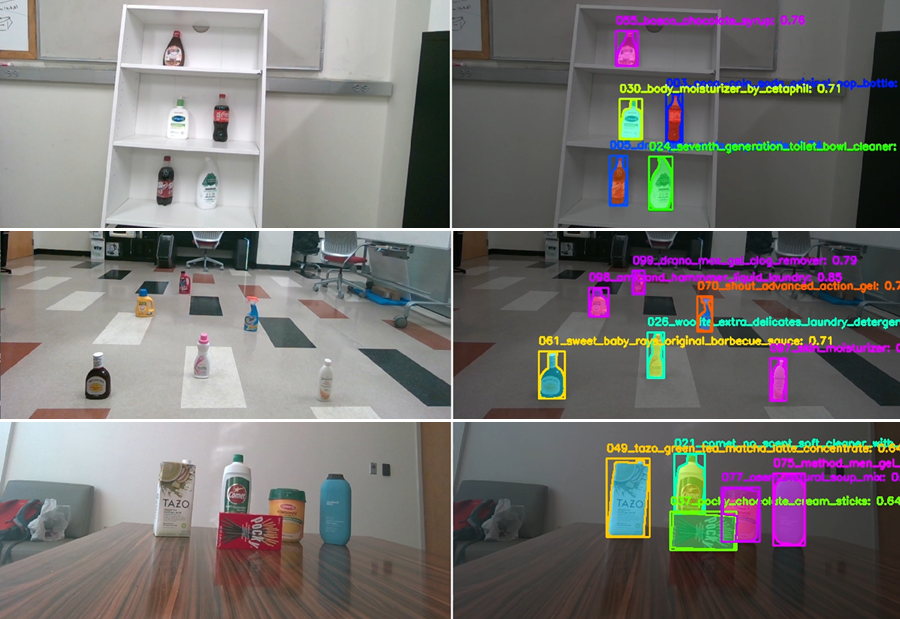}
    \caption{Evaluation of NIDS-Net on real-world objects across diverse scenes. All predictions are accurate in these examples.}
    \label{fig:real}
\end{figure}

\begin{figure}
    \centering
 \vspace{-3mm}    
\includegraphics[width=0.8\columnwidth]{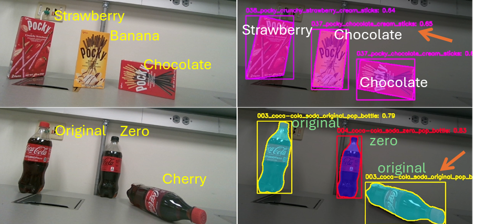}
\vspace{-2mm}
    \caption{Examples of failure cases from NIDS-Net, where the orange arrows indicate incorrect predictions.}
    \label{fig:failure}
    \vspace{-5mm}
\end{figure}

We capture 14 multi-view template images per object using seven Intel RealSense D455 cameras for real-world objects. With a trained weight adapter on these images, our method is tested on these objects with one D455 camera, demonstrating robust performance across various scenes (Fig. \ref{fig:real}). However, it misclassifies objects when highly similar instances are present as in Fig. \ref{fig:failure}. With the weight adapter trained on BOP datasets, we also tested our method on YCB-V objects using a Fetch robot. The demonstrations are included in the supplementary video.

\begin{table}[h]
\centering
\caption{The performance comparison of $\beta$ in the Weight Adapter on the high-resolution dataset. }
\label{tab:beta_test}
\vspace{-2mm}
{\scriptsize
\begin{tabular}{c|cccccc}
\hline
$\beta$ & 1 & 5 & 10 & 20 & 50 & 100 \\ \hline
AP & 62.0 & 63.1 & \textbf{63.9} & 63.1 & 62.7 & 62.2 \\
AP50 & 73.7 & 75.6 & \textbf{76.6} & 75.6 & 75.1 & 74.3 \\
AP75 & 68.3 & 69.5 & \textbf{70.6} & 69.3 & 69.0 & 68.2 \\ \hline
\end{tabular}
}
\vspace{-5mm}
\end{table}

\subsection{Ablation Study}
\textbf{Grounded-SAM (GS) vs SAM.} On the seven BOP datasets containing images of numerous cluttered scenes, we compare the object proposals from GS and SAM as detailed in Table \ref{tab:gsam}. The results indicate that GS yields more precise bounding boxes and masks compared to using SAM alone. Furthermore, GS enhances efficiency by eliminating false object proposals, thereby reducing runtime.

\textbf{FFA vs \(cls\) token.} To compare these two types of embedding generation, we present the results in Table \ref{tab:gsam}. Our weight adapter enhances both two types of embeddings. Despite close segmentation results, FFA produces embeddings that possess greater adaptive potential, demonstrated by higher AP scores after adaption via our weight adapter. 

\textbf{$\beta$ in our Weight Adapter}. We evaluate various values of $\beta$ on the high-resolution dataset. All results presented in Table \ref{tab:beta_test} surpass those obtained using CLIP-Adapter. We set $\beta$ to to 10 across all datasets as it yields the best results. Additional ablation studies are provided in the appendix.

\vspace{-1mm}
\section{Discussions}
\vspace{-1mm}
In this study, we introduce NIDS-Net, a framework designed for novel instance detection and segmentation. We utilize Grounding DINO and SAM to generate precise foreground object proposals, as opposed to conventional naive region proposals. We also introduce the Weight Adapter, which effectively refines features from a pre-trained vision model, mitigates the risk of overfitting, and adjusts the weighting of cosine similarities. With the adapter, template and proposal embeddings of different instances are separated to facilitate the subsequent matching. Our method surpasses other approaches significantly in detection performance and also excels in segmentation compared to existing methods. 

However, there are limitations. Given that our approach incorporates multiple pre-trained models, it requires greater computational resources compared to end-to-end detectors. When instances exhibit highly similar appearances, NIDS-Net may encounter detection failures. The method sometimes missed heavily occluded objects with low confidence scores. 

In this work, instances are represented by \(K\) template embeddings. For future research, we will explore using a single, distinctive embedding for each instance that acts as its identifier to enable one-shot detection. This approach will allow the detector to identify and locate a target instance within a query image using just one template image. Additionally, developing a computationally efficient method for this process will be crucial for its application in robotics.







\vspace{-1mm}
\section*{ACKNOWLEDGMENT}
\vspace{-1mm}
This work was supported in part by the DARPA
Perceptually-enabled Task Guidance (PTG) Program under contract number HR00112220005.

\bibliographystyle{IEEEtran}
\bibliography{reference}

\begin{thebibliography}{10}
\providecommand{\url}[1]{#1}
\csname url@samestyle\endcsname
\providecommand{\newblock}{\relax}
\providecommand{\bibinfo}[2]{#2}
\providecommand{\BIBentrySTDinterwordspacing}{\spaceskip=0pt\relax}
\providecommand{\BIBentryALTinterwordstretchfactor}{4}
\providecommand{\BIBentryALTinterwordspacing}{\spaceskip=\fontdimen2\font plus
\BIBentryALTinterwordstretchfactor\fontdimen3\font minus \fontdimen4\font\relax}
\providecommand{\BIBforeignlanguage}[2]{{%
\expandafter\ifx\csname l@#1\endcsname\relax
\typeout{** WARNING: IEEEtran.bst: No hyphenation pattern has been}%
\typeout{** loaded for the language `#1'. Using the pattern for}%
\typeout{** the default language instead.}%
\else
\language=\csname l@#1\endcsname
\fi
#2}}
\providecommand{\BIBdecl}{\relax}
\BIBdecl

\bibitem{li2024voxdet}
B.~Li, J.~Wang, Y.~Hu, C.~Wang, and S.~Scherer, ``Voxdet: Voxel learning for novel instance detection,'' \emph{Advances in Neural Information Processing Systems}, vol.~36, 2024.

\bibitem{shen2023instance}
Q.~Shen, Y.~Zhao, N.~Kwon, J.~Kim, Y.~Li, and S.~Kong, ``A high-resolution dataset for instance detection with multi-view instance capture,'' in \emph{NeurIPS Datasets and Benchmarks Track}, 2023.

\bibitem{nguyen2023cnos}
V.~N. Nguyen, T.~Groueix, G.~Ponimatkin, V.~Lepetit, and T.~Hodan, ``Cnos: A strong baseline for cad-based novel object segmentation,'' in \emph{Proceedings of the IEEE/CVF International Conference on Computer Vision}, 2023, pp. 2134--2140.

\bibitem{lin2023sam}
J.~Lin, L.~Liu, D.~Lu, and K.~Jia, ``Sam-6d: Segment anything model meets zero-shot 6d object pose estimation,'' \emph{arXiv:2311.15707}, 2023.

\bibitem{kirillov2023segment}
A.~Kirillov, E.~Mintun, N.~Ravi, H.~Mao, C.~Rolland, L.~Gustafson, T.~Xiao, S.~Whitehead, A.~C. Berg, W.-Y. Lo \emph{et~al.}, ``Segment anything,'' in \emph{Proceedings of the IEEE/CVF International Conference on Computer Vision}, 2023, pp. 4015--4026.

\bibitem{zhao2023fast}
X.~Zhao, W.~Ding, Y.~An, Y.~Du, T.~Yu, M.~Li, M.~Tang, and J.~Wang, ``Fast segment anything,'' \emph{arXiv preprint arXiv:2306.12156}, 2023.

\bibitem{oquab2023dinov2}
M.~Oquab, T.~Darcet, T.~Moutakanni, H.~Vo, M.~Szafraniec, V.~Khalidov, P.~Fernandez, D.~Haziza, F.~Massa, A.~El-Nouby \emph{et~al.}, ``Dinov2: Learning robust visual features without supervision,'' \emph{arXiv:2304.07193}, 2023.

\bibitem{darcet2023vision}
T.~Darcet, M.~Oquab, J.~Mairal, and P.~Bojanowski, ``Vision transformers need registers,'' \emph{arXiv preprint arXiv:2309.16588}, 2023.

\bibitem{liu2023groundingDINO}
S.~Liu, Z.~Zeng, T.~Ren, F.~Li, H.~Zhang, J.~Yang, C.~Li, J.~Yang, H.~Su, J.~Zhu \emph{et~al.}, ``Grounding dino: Marrying dino with grounded pre-training for open-set object detection,'' \emph{arXiv preprint arXiv:2303.05499}, 2023.

\bibitem{kotar2024these}
K.~Kotar, S.~Tian, H.-X. Yu, D.~Yamins, and J.~Wu, ``Are these the same apple? comparing images based on object intrinsics,'' \emph{Advances in Neural Information Processing Systems}, vol.~36, 2024.

\bibitem{oord2018representation}
A.~v.~d. Oord, Y.~Li, and O.~Vinyals, ``Representation learning with contrastive predictive coding,'' \emph{arXiv preprint arXiv:1807.03748}, 2018.

\bibitem{chen2020simple}
T.~Chen, S.~Kornblith, M.~Norouzi, and G.~Hinton, ``A simple framework for contrastive learning of visual representations,'' in \emph{International conference on machine learning}.\hskip 1em plus 0.5em minus 0.4em\relax PMLR, 2020, pp. 1597--1607.

\bibitem{gao2021clipAdapter}
P.~Gao, S.~Geng, R.~Zhang, T.~Ma, R.~Fang, Y.~Zhang, H.~Li, and Y.~Qiao, ``Clip-adapter: Better vision-language models with feature adapters,'' \emph{arXiv 2110.04544}, 2021.

\bibitem{mcvitie1971stable}
D.~G. McVitie and L.~B. Wilson, ``The stable marriage problem,'' \emph{Communications of the ACM}, vol.~14, no.~7, pp. 486--490, 1971.

\bibitem{sundermeyer2023bop}
M.~Sundermeyer, T.~Hoda{\v{n}}, Y.~Labbe, G.~Wang, E.~Brachmann, B.~Drost, C.~Rother, and J.~Matas, ``Bop challenge 2022 on detection, segmentation and pose estimation of specific rigid objects,'' in \emph{Proceedings of the IEEE/CVF Conference on Computer Vision and Pattern Recognition}, 2023, pp. 2784--2793.

\bibitem{dosovitskiy2020ViT}
A.~Dosovitskiy, L.~Beyer, A.~Kolesnikov, D.~Weissenborn, X.~Zhai, T.~Unterthiner, M.~Dehghani, M.~Minderer, G.~Heigold, S.~Gelly \emph{et~al.}, ``An image is worth 16x16 words: Transformers for image recognition at scale,'' \emph{arXiv:2010.11929}, 2020.

\bibitem{caron2021dino}
M.~Caron, H.~Touvron, I.~Misra, H.~J\'egou, J.~Mairal, P.~Bojanowski, and A.~Joulin, ``Emerging properties in self-supervised vision transformers,'' in \emph{Proceedings of the International Conference on Computer Vision (ICCV)}, 2021.

\bibitem{mercier2021deepTemplate}
J.-P. Mercier, M.~Garon, P.~Giguere, and J.-F. Lalonde, ``Deep template-based object instance detection,'' in \emph{Proceedings of the IEEE/CVF Winter Conference on Applications of Computer Vision}, 2021, pp. 1507--1516.

\bibitem{ammirato2018targetInstDet}
P.~Ammirato, C.-Y. Fu, M.~Shvets, J.~Kosecka, and A.~C. Berg, ``Target driven instance detection,'' \emph{arXiv preprint arXiv:1803.04610}, 2018.

\bibitem{liu2022gen6d}
Y.~Liu, Y.~Wen, S.~Peng, C.~Lin, X.~Long, T.~Komura, and W.~Wang, ``Gen6d: Generalizable model-free 6-dof object pose estimation from rgb images,'' in \emph{European Conference on Computer Vision}.\hskip 1em plus 0.5em minus 0.4em\relax Springer, 2022, pp. 298--315.

\bibitem{tip_adapter_eccv22}
R.~Zhang, Z.~Wei, R.~Fang, P.~Gao, K.~Li, J.~Dai, Y.~Qiao, and H.~Li, ``Tip-adapter: Training-free adaption of clip for few-shot classification,'' \emph{arXiv preprint arXiv:2207.09519}, 2022.

\bibitem{padalunkal2023protoclip}
J.~J. P, K.~Palanisamy, Y.-W. Chao, X.~Du, and Y.~Xiang, ``Proto-clip: Vision-language prototypical network for few-shot learning,'' 2023.

\bibitem{hu2018squeeze}
J.~Hu, L.~Shen, and G.~Sun, ``Squeeze-and-excitation networks,'' in \emph{Proceedings of the IEEE conference on computer vision and pattern recognition}, 2018, pp. 7132--7141.

\bibitem{ren2024grounded}
T.~Ren, S.~Liu, A.~Zeng, J.~Lin, K.~Li, H.~Cao, J.~Chen, X.~Huang, Y.~Chen, F.~Yan \emph{et~al.}, ``Grounded sam: Assembling open-world models for diverse visual tasks,'' \emph{arXiv:2401.14159}, 2024.

\bibitem{ren2015faster}
S.~Ren, K.~He, R.~Girshick, and J.~Sun, ``Faster r-cnn: Towards real-time object detection with region proposal networks,'' \emph{Advances in neural information processing systems}, vol.~28, 2015.

\bibitem{lin2017focal}
T.-Y. Lin, P.~Goyal, R.~Girshick, K.~He, and P.~Doll{\'a}r, ``Focal loss for dense object detection,'' in \emph{Proceedings of the IEEE international conference on computer vision}, 2017, pp. 2980--2988.

\bibitem{zhou2019objects}
X.~Zhou, D.~Wang, and P.~Kr{\"a}henb{\"u}hl, ``Objects as points,'' \emph{arXiv preprint arXiv:1904.07850}, 2019.

\bibitem{9010746}
Z.~Tian, C.~Shen, H.~Chen, and T.~He, ``Fcos: Fully convolutional one-stage object detection,'' in \emph{2019 IEEE/CVF International Conference on Computer Vision (ICCV)}, 2019, pp. 9626--9635.

\bibitem{zhang2022dinodetr}
H.~Zhang, F.~Li, S.~Liu, L.~Zhang, H.~Su, J.~Zhu, L.~M. Ni, and H.-Y. Shum, ``Dino: Detr with improved denoising anchor boxes for end-to-end object detection,'' 2022.

\bibitem{caron2021emerging}
M.~Caron, H.~Touvron, I.~Misra, H.~J{\'e}gou, J.~Mairal, P.~Bojanowski, and A.~Joulin, ``Emerging properties in self-supervised vision transformers,'' in \emph{Proceedings of the IEEE/CVF international conference on computer vision}, 2021, pp. 9650--9660.

\bibitem{kingma2014adam}
D.~P. Kingma and J.~Ba, ``Adam: A method for stochastic optimization,'' \emph{arXiv preprint arXiv:1412.6980}, 2014.

\bibitem{hinton2012improving}
G.~E. Hinton, N.~Srivastava, A.~Krizhevsky, I.~Sutskever, and R.~R. Salakhutdinov, ``Improving neural networks by preventing co-adaptation of feature detectors,'' \emph{arXiv preprint arXiv:1207.0580}, 2012.

\bibitem{brachmann2014learning}
E.~Brachmann, A.~Krull, F.~Michel, S.~Gumhold, J.~Shotton, and C.~Rother, ``Learning 6d object pose estimation using 3d object coordinates,'' in \emph{Computer Vision--ECCV 2014: 13th European Conference, Zurich, Switzerland, September 6-12, 2014, Proceedings, Part II 13}.\hskip 1em plus 0.5em minus 0.4em\relax Springer, 2014, pp. 536--551.

\bibitem{calli2015ycb}
B.~Calli, A.~Singh, A.~Walsman, S.~Srinivasa, P.~Abbeel, and A.~M. Dollar, ``The ycb object and model set: Towards common benchmarks for manipulation research,'' in \emph{International conference on advanced robotics (ICAR)}.\hskip 1em plus 0.5em minus 0.4em\relax IEEE, 2015, pp. 510--517.

\bibitem{kim2022learning}
D.~Kim, T.-Y. Lin, A.~Angelova, I.~S. Kweon, and W.~Kuo, ``Learning open-world object proposals without learning to classify,'' \emph{IEEE Robotics and Automation Letters}, vol.~7, no.~2, pp. 5453--5460, 2022.

\bibitem{radford2021learning}
A.~Radford, J.~W. Kim, C.~Hallacy, A.~Ramesh, G.~Goh, S.~Agarwal, G.~Sastry, A.~Askell, P.~Mishkin, J.~Clark \emph{et~al.}, ``Learning transferable visual models from natural language supervision,'' in \emph{International conference on machine learning}.\hskip 1em plus 0.5em minus 0.4em\relax PMLR, 2021, pp. 8748--8763.

\bibitem{osokin2020os2d}
A.~Osokin, D.~Sumin, and V.~Lomakin, ``Os2d: One-stage one-shot object detection by matching anchor features,'' in \emph{Computer Vision--ECCV 2020: 16th European Conference, Glasgow, UK, August 23--28, 2020, Proceedings, Part XV 16}.\hskip 1em plus 0.5em minus 0.4em\relax Springer, 2020, pp. 635--652.

\bibitem{yang2022balanced}
H.~Yang, S.~Cai, H.~Sheng, B.~Deng, J.~Huang, X.-S. Hua, Y.~Tang, and Y.~Zhang, ``Balanced and hierarchical relation learning for one-shot object detection,'' in \emph{Proceedings of the IEEE/CVF conference on computer vision and pattern recognition}, 2022, pp. 7591--7600.

\bibitem{lin2014microsoft}
T.-Y. Lin, M.~Maire, S.~Belongie, J.~Hays, P.~Perona, D.~Ramanan, P.~Doll{\'a}r, and C.~L. Zitnick, ``Microsoft coco: Common objects in context,'' in \emph{Computer Vision--ECCV 2014: 13th European Conference, Zurich, Switzerland, September 6-12, 2014, Proceedings, Part V 13}.\hskip 1em plus 0.5em minus 0.4em\relax Springer, 2014, pp. 740--755.

\bibitem{hodan2017t}
T.~Hodan, P.~Haluza, {\v{S}}.~Obdr{\v{z}}{\'a}lek, J.~Matas, M.~Lourakis, and X.~Zabulis, ``T-less: An rgb-d dataset for 6d pose estimation of texture-less objects,'' in \emph{2017 IEEE Winter Conference on Applications of Computer Vision (WACV)}.\hskip 1em plus 0.5em minus 0.4em\relax IEEE, 2017, pp. 880--888.

\bibitem{hodan2018bop}
T.~Hodan, F.~Michel, E.~Brachmann, W.~Kehl, A.~GlentBuch, D.~Kraft, B.~Drost, J.~Vidal, S.~Ihrke, X.~Zabulis \emph{et~al.}, ``Bop: Benchmark for 6d object pose estimation,'' in \emph{Proceedings of the European conference on computer vision (ECCV)}, 2018, pp. 19--34.

\bibitem{doumanoglou2016recovering}
A.~Doumanoglou, R.~Kouskouridas, S.~Malassiotis, and T.-K. Kim, ``Recovering 6d object pose and predicting next-best-view in the crowd,'' in \emph{Proceedings of the IEEE conference on computer vision and pattern recognition}, 2016, pp. 3583--3592.

\bibitem{drost2017introducing}
B.~Drost, M.~Ulrich, P.~Bergmann, P.~Hartinger, and C.~Steger, ``Introducing mvtec itodd-a dataset for 3d object recognition in industry,'' in \emph{Proceedings of the IEEE international conference on computer vision workshops}, 2017, pp. 2200--2208.

\bibitem{kaskman2019homebreweddb}
R.~Kaskman, S.~Zakharov, I.~Shugurov, and S.~Ilic, ``Homebreweddb: Rgb-d dataset for 6d pose estimation of 3d objects,'' in \emph{Proceedings of the IEEE/CVF International Conference on Computer Vision Workshops}, 2019, pp. 0--0.

\bibitem{denninger2019blenderproc}
M.~Denninger, M.~Sundermeyer, D.~Winkelbauer, Y.~Zidan, D.~Olefir, M.~Elbadrawy, A.~Lodhi, and H.~Katam, ``Blenderproc,'' \emph{arXiv preprint arXiv:1911.01911}, 2019.

\bibitem{chen20233d}
J.~Chen, M.~Sun, T.~Bao, R.~Zhao, L.~Wu, and Z.~He, ``3d model-based zero-shot pose estimation pipeline,'' \emph{arXiv:2305.17934}, 2023.

\bibitem{mercier2021deep}
J.-P. Mercier, M.~Garon, P.~Giguere, and J.-F. Lalonde, ``Deep template-based object instance detection,'' in \emph{Proceedings of the IEEE/CVF Winter Conference on Applications of Computer Vision}, 2021, pp. 1507--1516.

\bibitem{zhang2023faster}
C.~Zhang, D.~Han, Y.~Qiao, J.~U. Kim, S.-H. Bae, S.~Lee, and C.~S. Hong, ``Faster segment anything: Towards lightweight sam for mobile applications,'' \emph{arXiv:2306.14289}, 2023.

\bibitem{ke2024segment}
L.~Ke, M.~Ye, M.~Danelljan, Y.-W. Tai, C.-K. Tang, F.~Yu \emph{et~al.}, ``Segment anything in high quality,'' \emph{Advances in Neural Information Processing Systems}, vol.~36, 2024.

\bibitem{touvron2022deit}
H.~Touvron, M.~Cord, and H.~J{\'e}gou, ``Deit iii: Revenge of the vit,'' in \emph{European conference on computer vision}.\hskip 1em plus 0.5em minus 0.4em\relax Springer, 2022, pp. 516--533.

\end{thebibliography}

\clearpage
\section*{APPENDIX}
\section{Training Details}
\textbf{Detection.} 
For detection datasets, the weight adapter is trained with a batch size of 1024, while the CLIP-Adapter is trained with a batch size of 512 to enhance performance. Both adapters are trained for the same number of epochs, as detailed in Table \ref{tab:epochs}. To utilize Grounding DINO, a box threshold of 0.15 is set for the high-resolution and YCB-V datasets, and 0.10 for other datasets.

\begin{table}[ht]
\centering
\caption{The training epochs of different detection datasets.}
\label{tab:epochs}
{\scriptsize
\begin{tabular}{ccccc}
\hline
 & High-resolution & RoboTools & LM-O & YCB-V \\ \hline
Both Adapters & 40 & 80 & 40 & 40 \\ \hline
\end{tabular}}
\end{table}

\textbf{Segmentation.} We combine all instances from seven core datasets of the BOP benchmark. We train both adapters with a batch size of 1344 (32 instances \(\times\) 42 templates per instance) for 500 epochs. For Grounding DINO, a box threshold of 0.10 is set for all datasets.

\section{More Ablation Study}

\begin{table}
\caption{Detection results using different ViT backbones of Dinov2. ``reg'' indicates DINOv2 with registers. The results are based on all testing images of the high-resolution dataset. \textbf{WA Diff} indicates the improvement attributed to the Weight Adapter. }
\label{tab:dino_diff}
\vspace{-3mm}
\centering
{\scriptsize
\begin{tabular}{lcccc}
\hline
Dinov2 backbone & AP & AP50 & AP75 & \textbf{WA Diff} \\ \hline
ViT-S/14 & 47.5 & 56.4 & 51.7 & \multirow{2}{*}{+ 0.9 AP} \\
ViT-S/14 + WA & 48.4 & 57.5 & 52.9 &  \\ \hline
ViT-B/14 & 54.7 & 65.2 & 60.1 & \multirow{2}{*}{+ 3.2 AP} \\
ViT-B/14 + WA & 57.9 & 69.0 & 63.6 &  \\ \hline
ViT-L/14 & 56.8 & 67.7 & 62.3 & \multirow{2}{*}{+ 3.6 AP} \\
ViT-L/14 + WA & 60.4 & 71.8 & 66.4 &  \\ \hline
ViT-L/14 reg & 59.3 & 71.1 & 65.1 & \multirow{2}{*}{+ 4.6 AP} \\
ViT-L/14 reg +WA & 63.9 & 76.6 & 70.6 &  \\ \hline
\end{tabular}}
\vspace{-2mm}
\end{table}

\begin{table}
\caption{Detection results on the High-resolution real-world detection dataset using various SAM variants.}
\label{tab:sam_diff}
\vspace{-3mm}
\centering
{\scriptsize
\begin{tabular}{lcccc}
\hline
SAM Variant & AP & AP50 & AP75 & Time (sec) \\ \hline
Mobile SAM (Tiny ViT)~\cite{zhang2023faster} & 54.5 & 70.5 & 62.0 & 6.80 \\
Mobile SAM (Tiny ViT)~\cite{zhang2023faster} + WA & 58.2 & 75.0 & 66.5 & \textbf{6.73} \\ \hline
HQ-SAM (ViT-H)~\cite{ke2024segment} & 59.8 & 71.5 & 65.7 & 7.38 \\
HQ-SAM (ViT H)~\cite{ke2024segment} + WA & 63.8 & 76.2 & 70.5 & 7.31 \\ \hline
SAM (ViT-H) (Ours)~\cite{kirillov2023segment} & 59.3 & 71.1 & 65.1 & 6.92 \\
SAM (ViT-H) + WA (Ours)~\cite{kirillov2023segment} & \textbf{63.9} & \textbf{76.6} & \textbf{70.6} & 6.78 \\ \hline
\end{tabular}
}
\vspace{-2mm}
\end{table}

\begin{table}[ht]
\centering
\caption{The detection results on all images of the high-resolution dataset.  ``Proposal'' refers to the object proposal method. ``Embedding'' denotes the method of instance embedding generation.}
\label{tab:det_cls}
{\scriptsize
\begin{tabular}{ccccc}
\hline
Proposal & Embedding & AP & AP50 & AP75 \\ \hline
SAM & \(cls\) token & 41.6 & 49.1 & 46.0 \\
GS & \(cls\) token & 54.9 & 65.4 & 60.1 \\
GS & FFA & \textbf{59.3} & \textbf{71.1} & \textbf{65.1} \\ \hline
\end{tabular}}
\end{table}

\begin{figure*}
    \centering
\includegraphics[width=1.8\columnwidth]{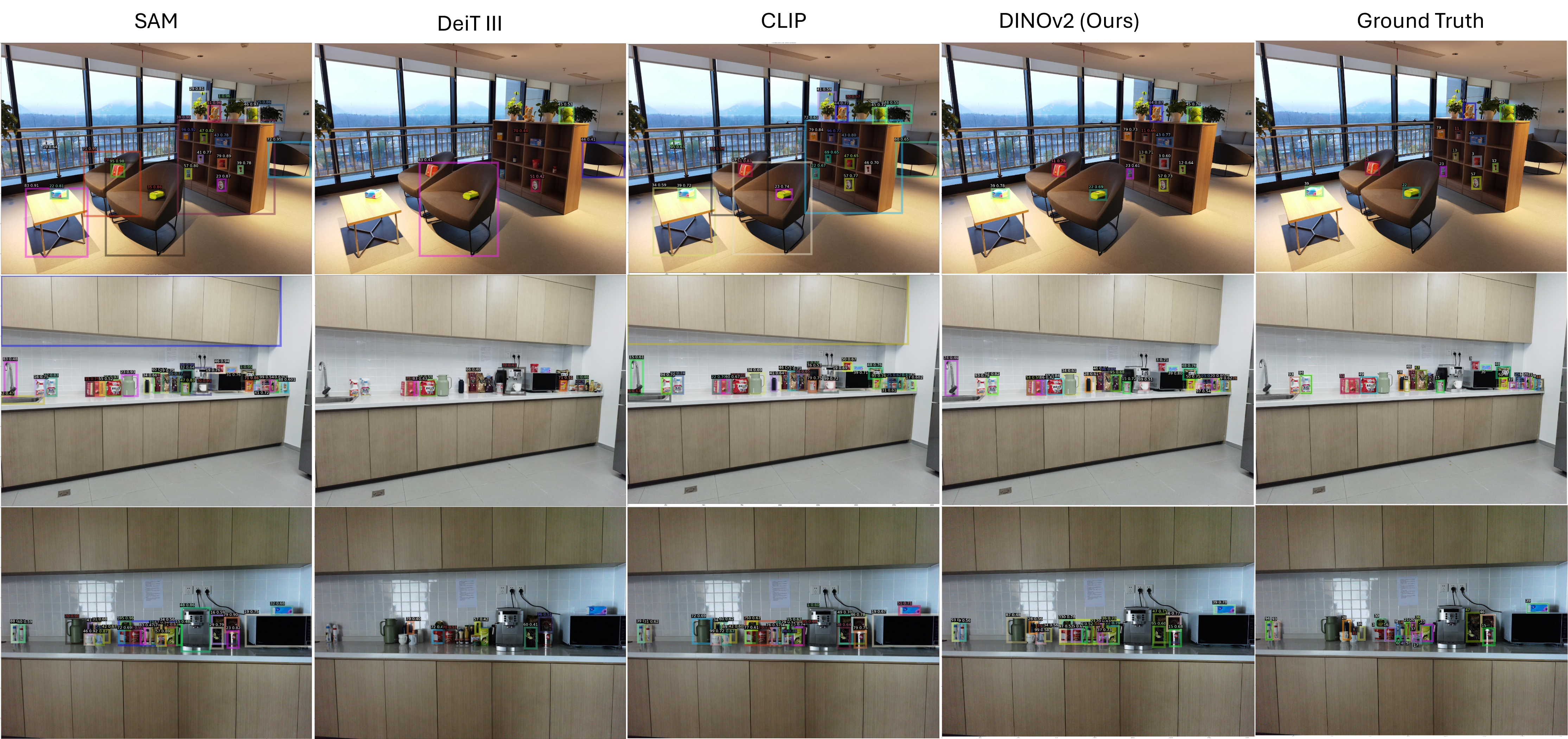}
    \caption{Visual detection results on the High-resolution dataset using different image encoders. }
    \label{fig:encoders}
\end{figure*}

\textbf{Image encoder.} Given the same object proposals with GS, we evaluate the FFA embeddings from different image encoders on the High-resolution dataset~\cite{shen2023instance}. As illustrated in Table \ref{tab:encoders}, DINOv2 exhibits superior performance attributable to its robust visual features. Fig. \ref{fig:encoders} presents the visual results of various image encoders.

\begin{table*}[ht]
\centering
\caption{The instance embeddings of different image encoders.}
\label{tab:encoders}
{\scriptsize
\begin{tabular}{lcccc}
\hline
Image Encoder & Feature Dimension & AP & AP50 & AP75 \\ \hline
SAM (Vit-L)~\cite{kirillov2023segment} & 256 & 0.9 & 1.1 & 0.9 \\
SAM (Vit-L)~\cite{kirillov2023segment} + WA & 256 & 0.7 & 0.8 & 0.7 \\ \hline
DeiT III~\cite{touvron2022deit} (Vit-L) & 784 & 2.9 & 3.6 & 3.1 \\
DeiT III~\cite{touvron2022deit} (Vit-L) + WA & 784 & 2.8 & 3.4 & 3.0 \\ \hline
CLIP (Vit-L)~\cite{ radford2021learning} & 1024 & 17.9 & 21.1 & 18.9 \\
CLIP (Vit-L)~\cite{ radford2021learning} + WA & 1024 & 20.4 & 24.0 & 21.8 \\ \hline
DINOv2 (Vit-L)~\cite{oquab2023dinov2}  & 1024 & 56.8 & 67.7 & 62.3 \\
DINOv2 (Vit-L)~\cite{oquab2023dinov2}  + WA & 1024 & 60.4 & 71.8 & 66.4 \\ \hline
DINOv2 (Vit-L reg)~\cite{oquab2023dinov2}  & 1024 & 59.3 & 71.1 & 65.1 \\
DINOv2 (Vit-L reg)~\cite{oquab2023dinov2}  + WA & 1024 & \textbf{63.9} & \textbf{76.6} & \textbf{70.6} \\ \hline
\end{tabular}}
\end{table*}

\textbf{Dinov2 backbones with adapter.} Our Weight Adapter is compatible with various backbones of Dinov2. Notably, more powerful backbones, which offer a more effective feature space, enable our adapter to deliver greater improvements. Details of this comparison are provided in Table \ref{tab:dino_diff}. 

\textbf{Aggregation.} For \(\text{cls}\) token embeddings, averaging the top \(k\) highest (\textit{avg k}) scores yields the best results~\cite{nguyen2023cnos, lin2023sam}. For FFA embeddings, the \(\max\) aggregation function achieves optimal outcomes. The comparison of these aggregation functions is detailed in Table \ref{tab:agg}.

\textbf{SAM variants.} We evaluate different SAM variants on the High-resolution real world detection dataset. As illustrated in Table \ref{tab:sam_diff}, SAM and our weight adapter achieves the superior performance. Given the precise bounding boxes, HQ-SAM\cite{ke2024segment} yields results comparable to those of SAM~\cite{kirillov2023segment}.

\begin{table*}[ht]
\centering
\caption{Comparison of aggregation functions for segmentation performance. We report Average Precision (AP). \textit{avg k} refers to averaging the top \(k\) scores. All results are based on object proposals from Grounded SAM (GS).}
\label{tab:agg}
{\scriptsize
\begin{tabular}{cccccccccc}
\hline
\multirow{2}{*}{Embedding} & \multirow{2}{*}{Aggregation} & \multicolumn{7}{c}{BOP Datasets} &  \\ \cline{3-9}
 &  & LM-O & T-LESS & TUD-L & IC-BIN & ITODD & HB & YCB-V & Mean \\ \hline
 \(cls\) token & \textit{avg k} & 41.7 & \textbf{41.7} & \textbf{50.8} & \textbf{31.5} & \textbf{30} & \textbf{58} & \textbf{63} & \textbf{45.2} \\
\(cls\) token & \(\max\) & \textbf{42} & 37.4 & 45.9 & 30.2 & 28.9 & 54.9 & 61.5 & 43.0 \\
 \hline
FFA & \textit{avg k} & 42.5 & 42 & 47.4 & 28.2 & 27.3 & 55.1 & \textbf{61.5} & 43.4 \\
FFA & \(\max\) & \textbf{42.9} & \textbf{43} & \textbf{52} & \textbf{30.5} & \textbf{28.8} & \textbf{56.6} & 59.7 & \textbf{44.8} \\ \hline
\end{tabular}
}
\end{table*}

\textbf{Runtime.} We compare the efficiency of existing methods for novel instance segmentation, as presented in Table \ref{tab:bop_time}. Our approach significantly reduces running time by proposing only high-quality bounding boxes.

\begin{table*}[ht]
\centering
\caption{Runtime comparisons of various methods for novel instance segmentation.}
\label{tab:bop_time}
{\scriptsize
\begin{tabular}{lccc}
\hline
Method & Proposal & Server & \multicolumn{1}{l}{Time (sec)} \\ \hline
CNOS~\cite{nguyen2023cnos} & \multirow{4}{*}{FastSAM} & Tesla V100 & 0.22 \\
CNOS~\cite{nguyen2023cnos} &  & DeForce RTX 3090 & 0.23 \\
SAM-6D (RGB)~\cite{lin2023sam} &  & DeForce RTX 3090 & 0.25 \\
SAM-6D (RGBD)~\cite{lin2023sam} &  & DeForce RTX 3090 & 0.45 \\ \hline
CNOS~\cite{nguyen2023cnos} & \multirow{4}{*}{SAM} & Tesla V100 & 1.84 \\
CNOS~\cite{nguyen2023cnos} &  & DeForce RTX 3090 & 2.35 \\
SAM-6D (RGB)~\cite{lin2023sam} &  & DeForce RTX 3090 & 2.28 \\
SAM-6D (RGBD)~\cite{lin2023sam} &  & DeForce RTX 3090 & 2.80 \\ \hline
NIDS-Net w/o adapter (Ours) & \multirow{4}{*}{GS} & \multirow{4}{*}{RTX A5000} & 0.49 \\
NIDS-Net + CA (Ours) &  &  & 0.48 \\
NIDS-Net + WA (Ours) &  &  & 0.48 \\
NIDS-Net + WA + \(s_{\text{appe}}\) (Ours) &  &  & 0.48 \\ \hline
\end{tabular}}
\end{table*}

\section{Unseen Detection of BOP Benchmark}
 We compare our approach with ZeroPose~\cite{chen20233d}, CNOS~\cite{nguyen2023cnos}, and SAM-6D~\cite{lin2023sam} for 2D unseen detection, as illustrated in Table \ref{tab:bop_det}. Our method outperforms the best RGB method by 2.5 AP and competes effectively with the top RGB-D method.

\begin{table*}[ht]
\centering
\caption{Unseen instance detection results across the seven core datasets of the BOP benchmark, with all results reported as Average Precision (AP).}
\label{tab:bop_det}
{\scriptsize
\begin{tabular}{lccccccccc}
\hline
\multirow{2}{*}{Method} & \multirow{2}{*}{Proposal} & \multicolumn{7}{c}{BOP Datasets} &  \\ \cline{3-9}
 &  & LM-O & T-LESS & TUD-L & IC-BIN & ITODD & HB & YCB-V & Mean \\ \hline
ZeroPose~\cite{chen20233d} & SAM & 36.7 & 30.0 & 43.1 & 22.8 & 25.0 & 39.8 & 41.6 & 34.1 \\
CNOS~\cite{nguyen2023cnos}   & SAM & 39.5 & 33.0 & 36.8 & 20.7 & 31.3 & 42.3 & 49.0 & 36.1 \\
CNOS~\cite{nguyen2023cnos}  & FastSAM & 43.3 & 39.5 & 53.4 & 22.6 & 32.5 & 51.7 & 56.8 & 42.8 \\
SAM-6D (RGB)~\cite{lin2023sam} & FastSAM & 43.8 & 41.7 & 54.6 & 23.4 & 37.4 &  52.3 & 57.3 & 44.4 \\
SAM-6D (RGBD)~\cite{lin2023sam} & SAM & \textbf{46.6} & 43.7 & 53.7 & \textbf{26.1} & 39.3 & 53.1 & 51.9 & 44.9 \\
SAM-6D (RGBD)~\cite{lin2023sam} & FastSAM & 46.3 & 45.8 & 57.3 & 24.5 & \textbf{41.9} & 55.1 & 58.9 & \textbf{47.1} \\ \hline
NIDS-Net w/o adapter (Ours) & \multirow{3}{*}{GS} & 44.9 & 42.8 & 43.4 & 24.4 & 34.9 & 54.8 & 56.5 & 43.1 \\
NIDS-Net + WA (Ours) &  & 44.9 & 48.9 & 46.0 & 24.5 & 36.0 & \textbf{59.4} & \textbf{62.4} & 46.0 \\
NIDS-Net + WA + \(s_{\text{appe}}\) (Ours) &  & 45.7 & \textbf{49.3} & \textbf{48.6} & 25.7 & 37.9 & 58.7 & 62.1 & 46.9 \\ \hline
\end{tabular}
}
\end{table*}

\section{More Qualitative Results}
\subsection{Adapter}
\begin{figure*}
    \centering
\includegraphics[width=1.6\columnwidth]{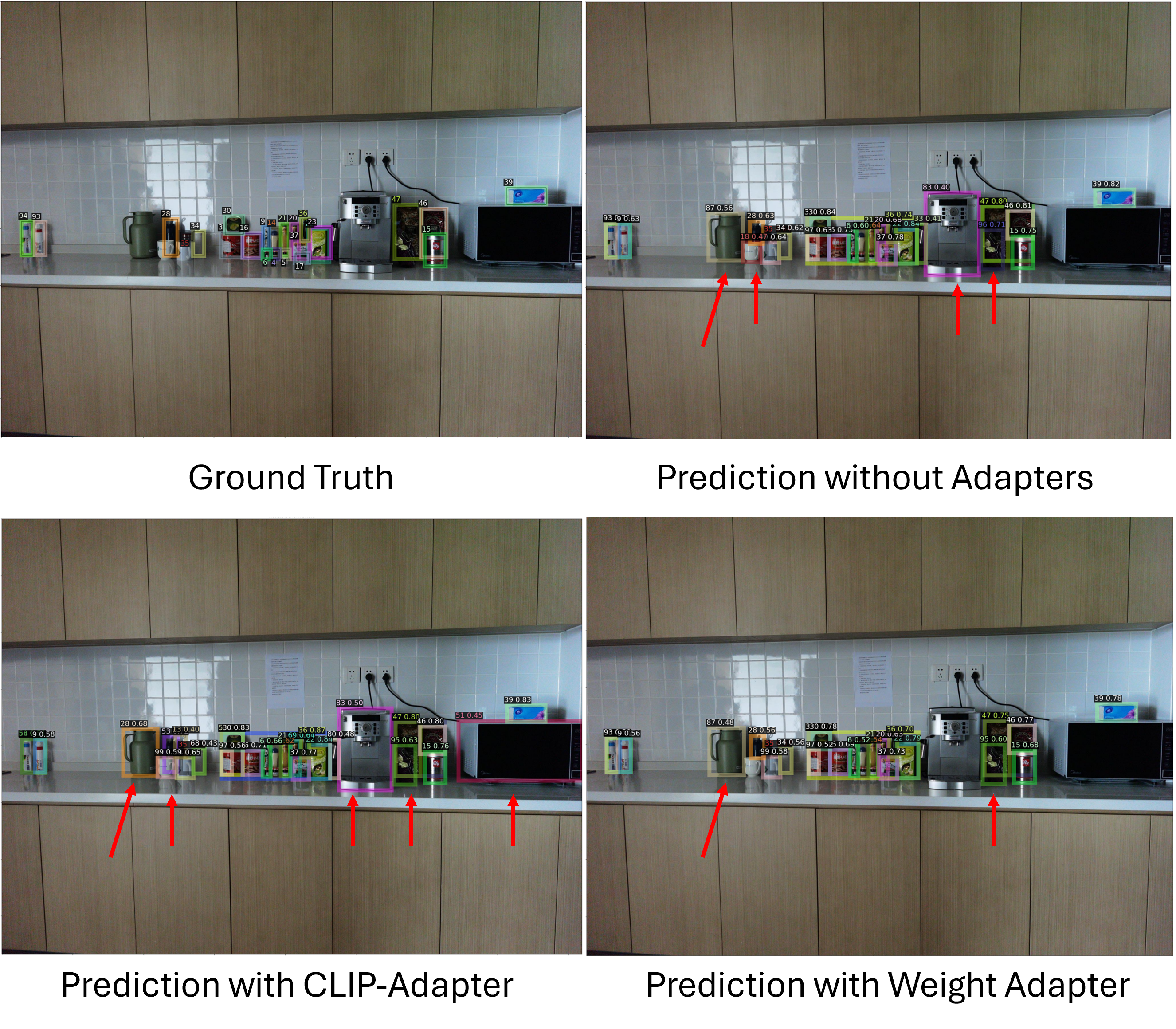}
    \caption{Comparison of different adapters on a hard scene image from the high-resolution dataset. Red arrows denote non-target objects that are erroneously classified as targets.}
    \label{fig:adapters_comparison}
\end{figure*}

To facilitate a comparison between the CLIP-Adapter and our Weight Adapter, we present a visual illustration in Figure \ref{fig:adapters_comparison}. It is evident that the CLIP-Adapter alters the feature space and spoils the embeddings of non-target objects due to overfitting. In contrast, our Weight Adapter delivers robust embeddings within the original feature space.

\subsection{Detection}
\begin{figure*}
    \centering
\includegraphics[width=1.0\columnwidth]{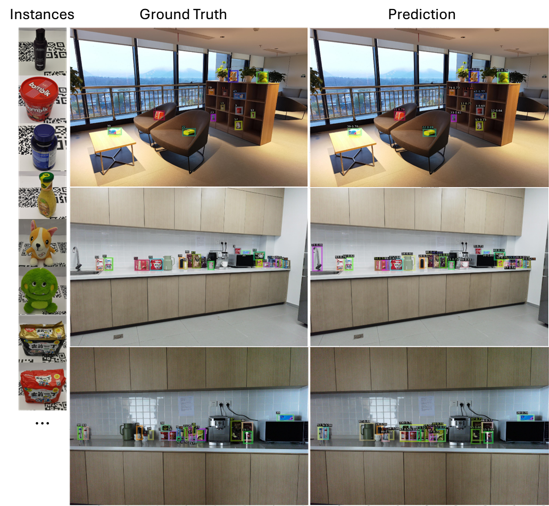} 
    \caption{Visual examples of our results using the Weight Adapter on the high-resolution dataset. Our approach detects specific object instances in cluttered scenes according to their real template images.}
    \label{fig:HR}
\end{figure*}

\begin{figure*}
    \centering
\includegraphics[width=1.6\columnwidth]{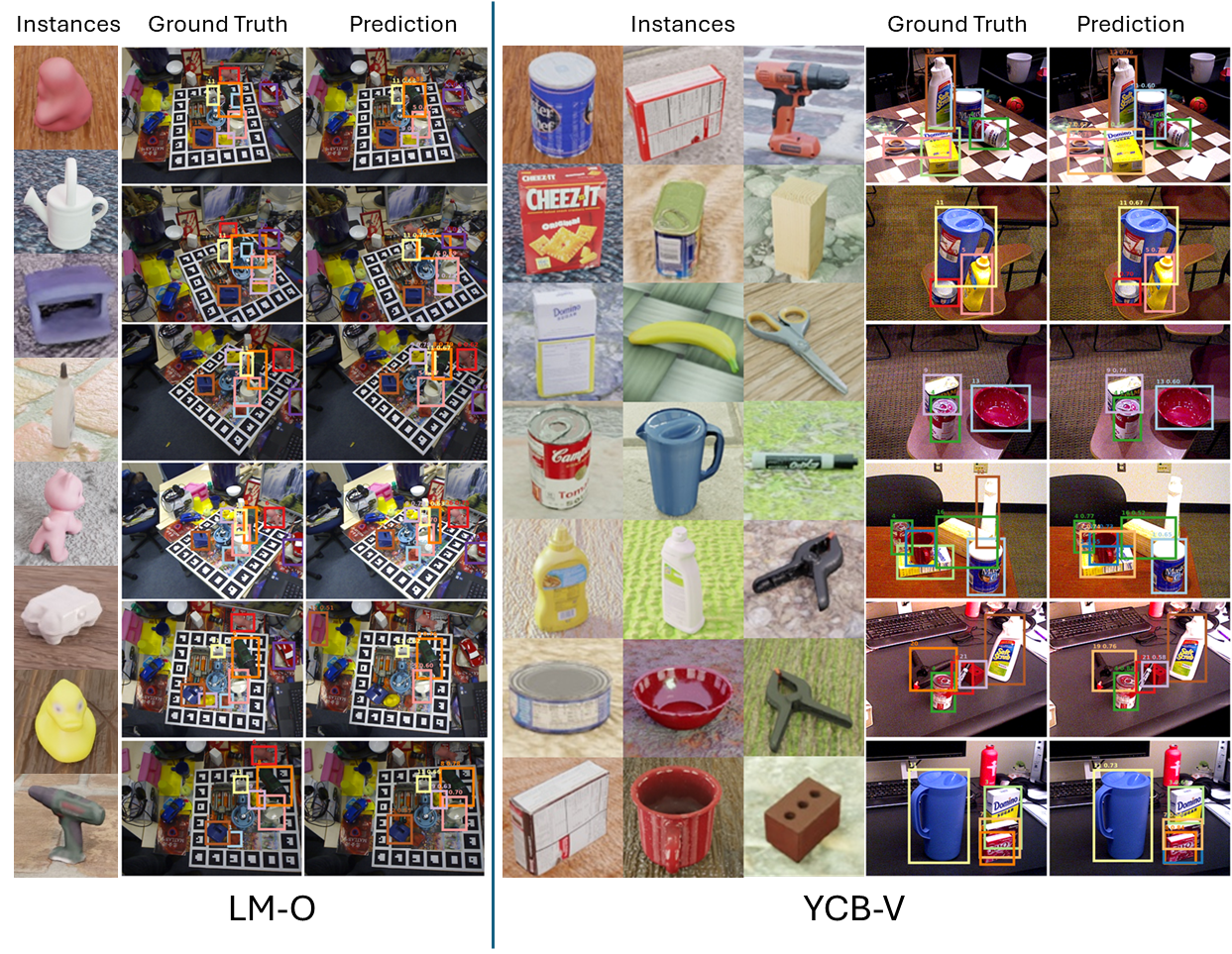}
    \caption{Visual detection results using the Weight Adapter on the LM-O and YCB-V datasets. Our approach detects object instances according to their synthetic template images.}
    \label{fig:lmo}
\end{figure*}

We display the visual outcomes of our methodology with the weight adapter on the LMO and YCB-V datasets in Figure \ref{fig:lmo}. The gap between synthetic and real images results in some instances of detection failure. For example, some LM-O instances are not found with our method. The examples of the high-resolution dataset are presented in Fig. \ref{fig:HR}.

\subsection{Segmentation}
\begin{figure*}
    \centering
\includegraphics[width=1.6\columnwidth]{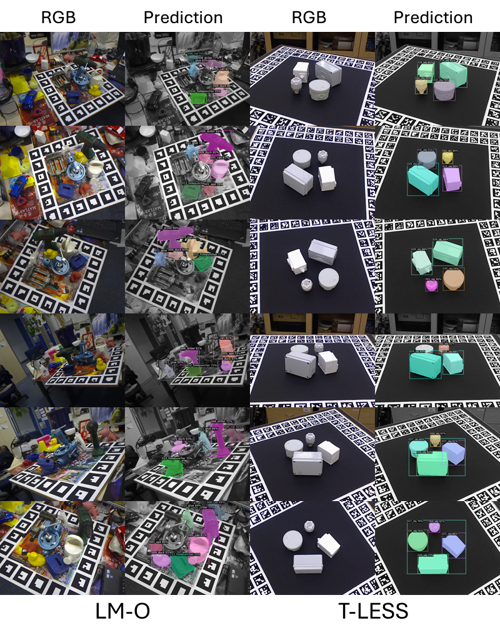} 
    \caption{Qualitative segmentation results on the LM-O and T-Less datasets.}
    \label{fig:seg1}
\end{figure*}

\begin{figure*}
    \centering
\includegraphics[width=1.6\columnwidth]{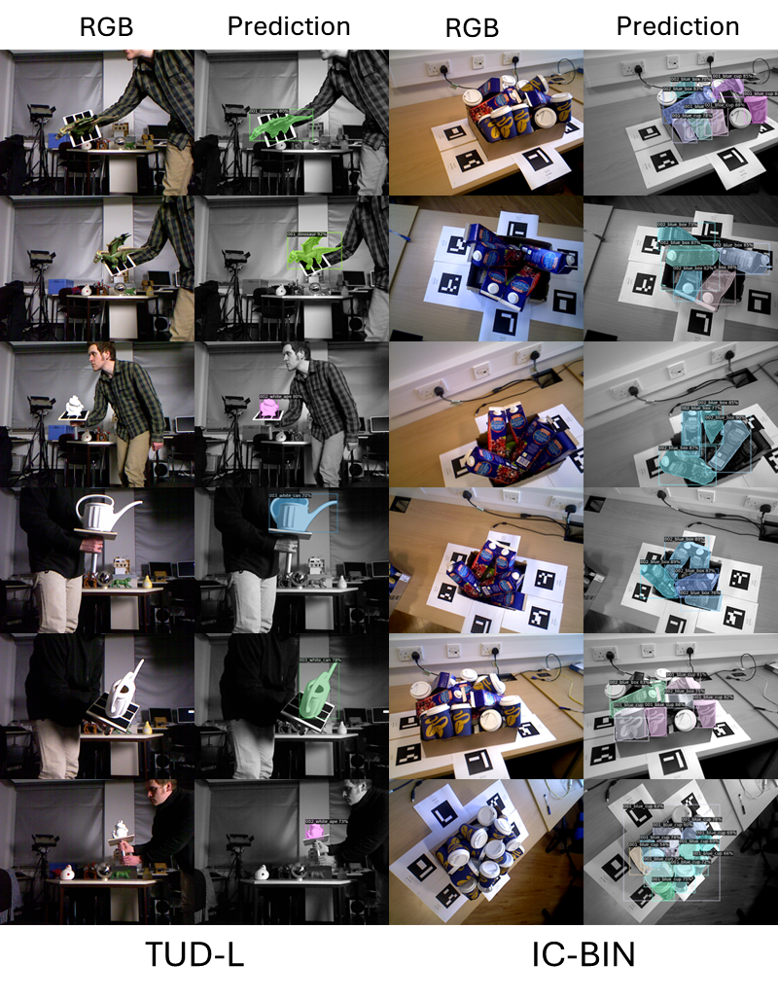}  
    \caption{Qualitative segmentation results on the TUD-L and IC-BIN datasets.}
    \label{fig:seg2}
\end{figure*}

\begin{figure*}
    \centering
\includegraphics[width=1.6\columnwidth]{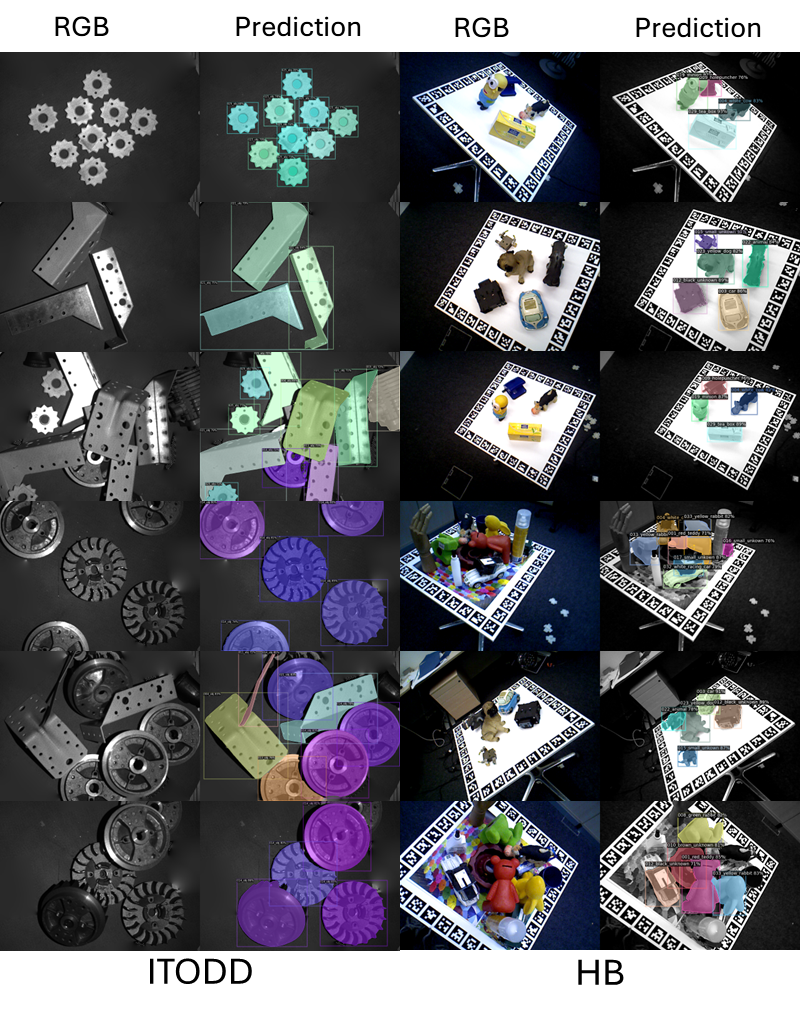} 
    \caption{Qualitative segmentation results on the ITODD and HB datasets}
    \label{fig:seg3}
\end{figure*}

\begin{figure*}
    \centering
\includegraphics[width=1.6\columnwidth]{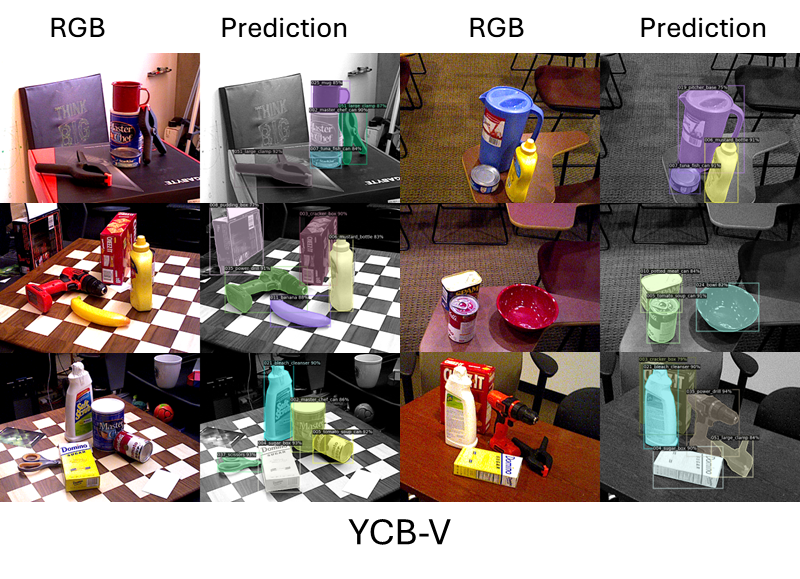}
    \caption{Qualitative segmentation results on the YCB-V dataset.}
    \label{fig:seg4}
\end{figure*}

We present the visual results of our approach using the weight adapter on the BoP datasets in Figures \ref{fig:seg1}, \ref{fig:seg2}, \ref{fig:seg3}, and \ref{fig:seg4}. These images demonstrate the effectiveness of our approach in cluttered environments. In some cases of T-LESS and IC-BIN datasets, Grounding DINO generates large bounding boxes which include multiple objects, causing under-segmentation. Furthermore, in IC-BIN and HB datasets, some heavily occluded objects with low confidence scores are overlooked by our method.


\end{document}